%% file: main.tex
\documentclass[journal]{IEEEtran}
\usepackage{cite}
\usepackage[pdftex]{graphicx}
\usepackage{amsmath}
\usepackage{array}
\usepackage[para,online,flushleft]{threeparttable}
\usepackage{url}
\usepackage{color}
\usepackage{subfigure}
\usepackage{makecell}
\usepackage{amssymb}
\usepackage{multirow}

\usepackage{color}
\definecolor{darkspringgreen}{rgb}{0.09, 0.45, 0.27}

\usepackage[normalem]{ulem}

\newcommand{\revRemove}[1]{}
\newcommand{\revFloatRemove}[1]{}
\newcommand{\revEqRemove}[1]{}
\newcommand{\revtwo}[1]{}
\newcommand{\revtwoRemove}[1]{}

\usepackage{tikz}
\usepackage{textcomp}
\usepackage[doipre={DOI:~}]{uri}
\usepackage{lipsum}
\newcommand\copyrighttext{%
  \footnotesize \textcopyright 2023 IEEE. Personal use is permitted, but republication/redistribution requires IEEE permission. See https://www.ieee.org/publications/rights/index.html for more information.
  }
\newcommand{\copyrightnotice}{%
\begin{tikzpicture}[remember picture,overlay]
\node[anchor=south,yshift=10pt] at (current page.south) {\fbox{\parbox{\dimexpr\textwidth-\fboxsep-\fboxrule\relax}{\copyrighttext}}};
\end{tikzpicture}%
}

\usepackage{fancyhdr}

\begin{document}

\bstctlcite{IEEEexample:BSTcontrol}

\title{Efficient Deep Learning Models for Privacy-preserving People Counting on Low-resolution Infrared Arrays}

\author{Chen~Xie,~\IEEEmembership{Member,~IEEE,}
        Francesco~Daghero,~\IEEEmembership{Member,~IEEE,}
        Yukai~Chen,~\IEEEmembership{Member,~IEEE,}
        Marco Castellano,
        Luca Gandolfi,
        Andrea Calimera,~\IEEEmembership{Member,~IEEE,}
        Enrico Macii,~\IEEEmembership{Fellow,~IEEE,}
        Massimo Poncino,~\IEEEmembership{Fellow,~IEEE,}
        and~Daniele Jahier Pagliari~\IEEEmembership{Member,~IEEE}
\thanks{C. Xie, F. Daghero, A. Calimera, M. Poncino and D. Jahier Pagliari are with the Department
of Control and Computer Engineering, Politecnico di Torino, Turin, 10129, Italy, e-mail: name.first\_surname@polito.it.}
\thanks{Y. Chen is with IMEC, Leuven, 3001, Belgium, e-mail: yukai.chen@imec.be.}
\thanks{M. Castellano and L. Gandolfi are with ST Microelectronics S.r.l., Cornaredo, 20010, Italy, e-mail:name.surname@st.com.}
\thanks{E. Macii is with the Interuniversity Department of Regional and Urban Studies and Planning, Politecnico di Torino, Turin, 10129, Italy, e-mail:enrico.macii@polito.it}
\thanks{Manuscript received January XX, XXXX; revised January XX, XXXX.}

\thanks{Copyright (c) 20xx IEEE. Personal use of this material is permitted. However, permission to use this material for any other purposes must be obtained from the IEEE by sending a request to pubs-permissions@ieee.org.}

}

\markboth{Internet of Things Journal,~Vol.~X, No.~X, January~XXXX}%
{Xie \MakeLowercase{\textit{et al.}}: Efficient Deep Learning Models}

\maketitle
\copyrightnotice

\thispagestyle{fancy}
\fancyhead{}
\fancyhead[C]{This article has been accepted for publication in IEEE Internet of Things Journal. This is the author's version which has not been fully edited and content may change prior to final publication. Citation information: DOI 10.1109/JIOT.2023.3263290}

\begin{abstract}
\input{sec/abstract.tex}
\end{abstract}


\IEEEpeerreviewmaketitle

\section{Introduction}\label{sec:introduction}
\input{sec/introduction.tex}

\section{Background and Related Works}\label{sec:background}
\input{sec/background.tex}

\section{Materials and Methods}\label{sec:methods}
\input{sec/methods.tex}

\section{Experimental Results}\label{sec:results}
\input{sec/results.tex}

\section{Conclusion}\label{sec:conclusions}
\input{sec/conclusion.tex}

\section*{Acknowledgment}
This work has received funding from the ECSEL Joint Undertaking (JU) under grant agreement No 101007321. The JU receives support from the European Union’s Horizon 2020 research and innovation programme and France, Belgium, Czech Republic, Germany, Italy, Sweden, Switzerland, Turkey.

\ifCLASSOPTIONcaptionsoff
  \newpage
\fi

\bibliographystyle{IEEEtran}
\bibliography{IEEEabrv,./bib/library}

\begin{IEEEbiography}[{\includegraphics[width=1in,height=1.25in,clip,keepaspectratio]{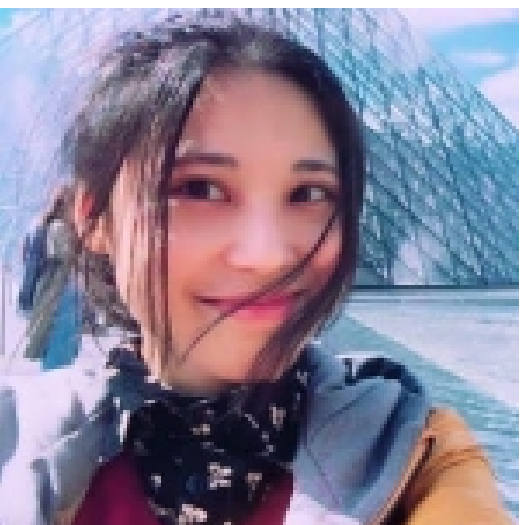}}]{Chen Xie} received the M.Sc degrees in Electronics Engineering at Politecnico di Torino in 2020. Since May 2020, she joined the EDA group in the Department of Control and Computer Engineering at Politecnico di Torino. Her main research interests concern energy-efficient implementations of machine learning algorithms and synthesis of smart sensors.
\end{IEEEbiography}

\begin{IEEEbiography}[{\includegraphics[width=1in,height=1.25in,clip,keepaspectratio]{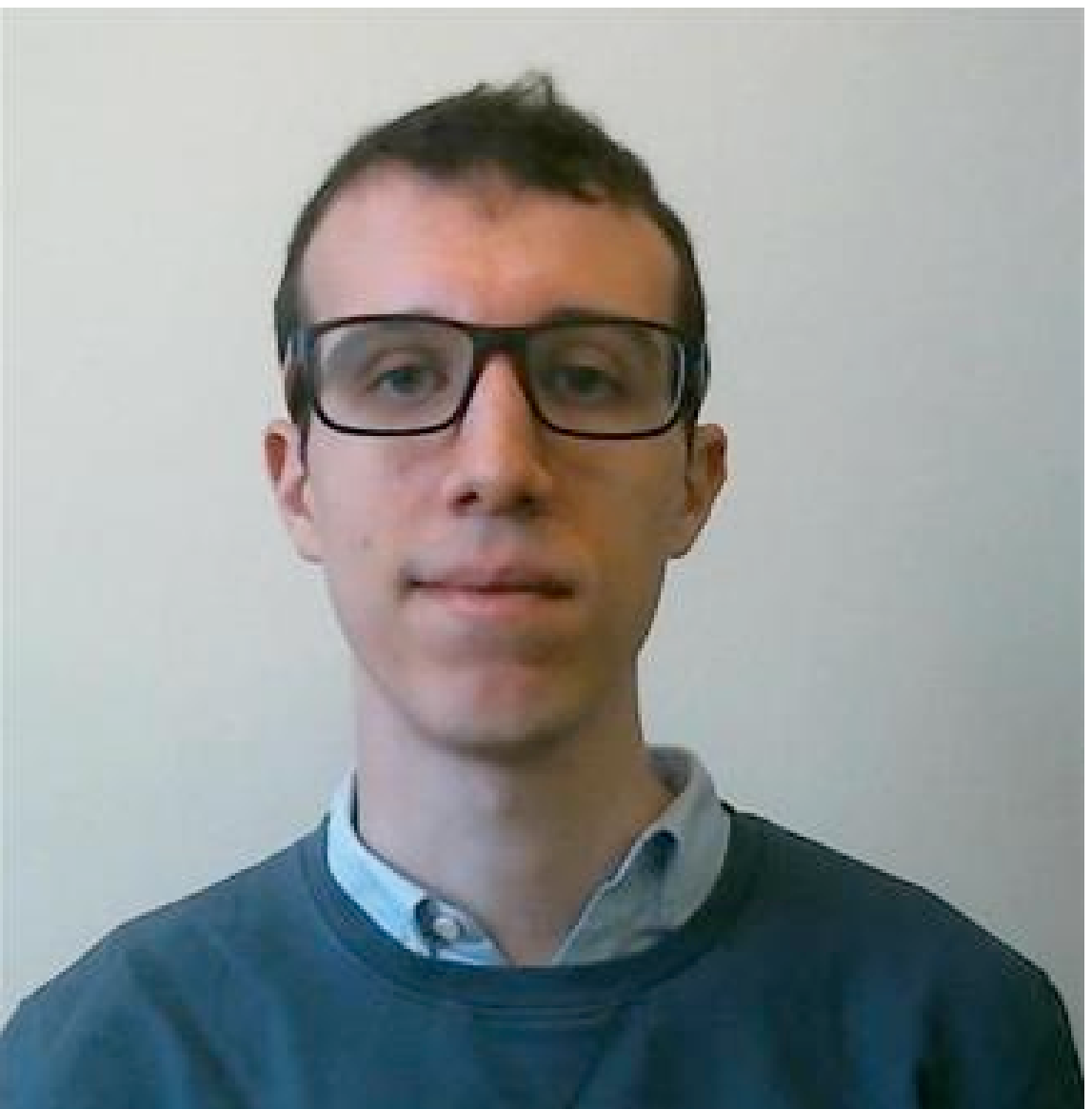}}]{Francesco Daghero} is a PhD student at Politecnico di Torino. He received a M.Sc. degree in computer engineering from Politecnico di Torino, Italy, in 2019. His research interests concern embedded machine learning and Industry 4.0.
\end{IEEEbiography}

\begin{IEEEbiography}[{\includegraphics[width=1in,height=1.25in,clip,keepaspectratio]{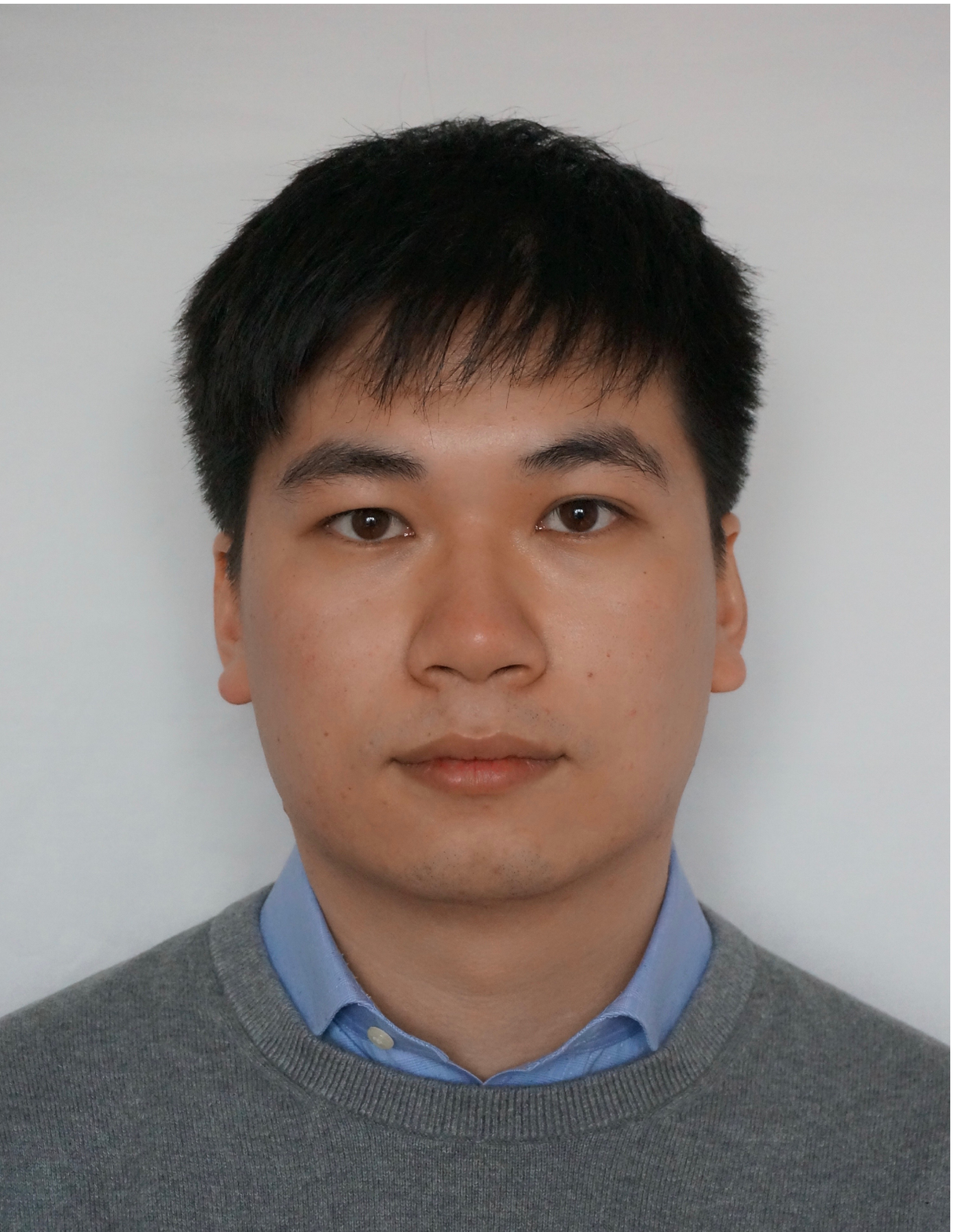}}]
{Yukai Chen} earned his M.Sc. and Ph.D. degrees in Computer Engineering from the Politecnico di Torino, Turin, Italy, in 2014 and 2018, respectively. He currently serves as a Senior Researcher at IMEC, where he contributes to the System and Technology Co-optimization Program. His primary focus is on system-level power and thermal management for High-Performance Computing Architectures. His research interests encompass design automation for non-functional property modeling, simulation, and optimization, with particular emphasis on energy-efficient design and design space exploration.
\end{IEEEbiography}

\begin{IEEEbiography}[{\includegraphics[width=1in,height=1.25in,clip,keepaspectratio]{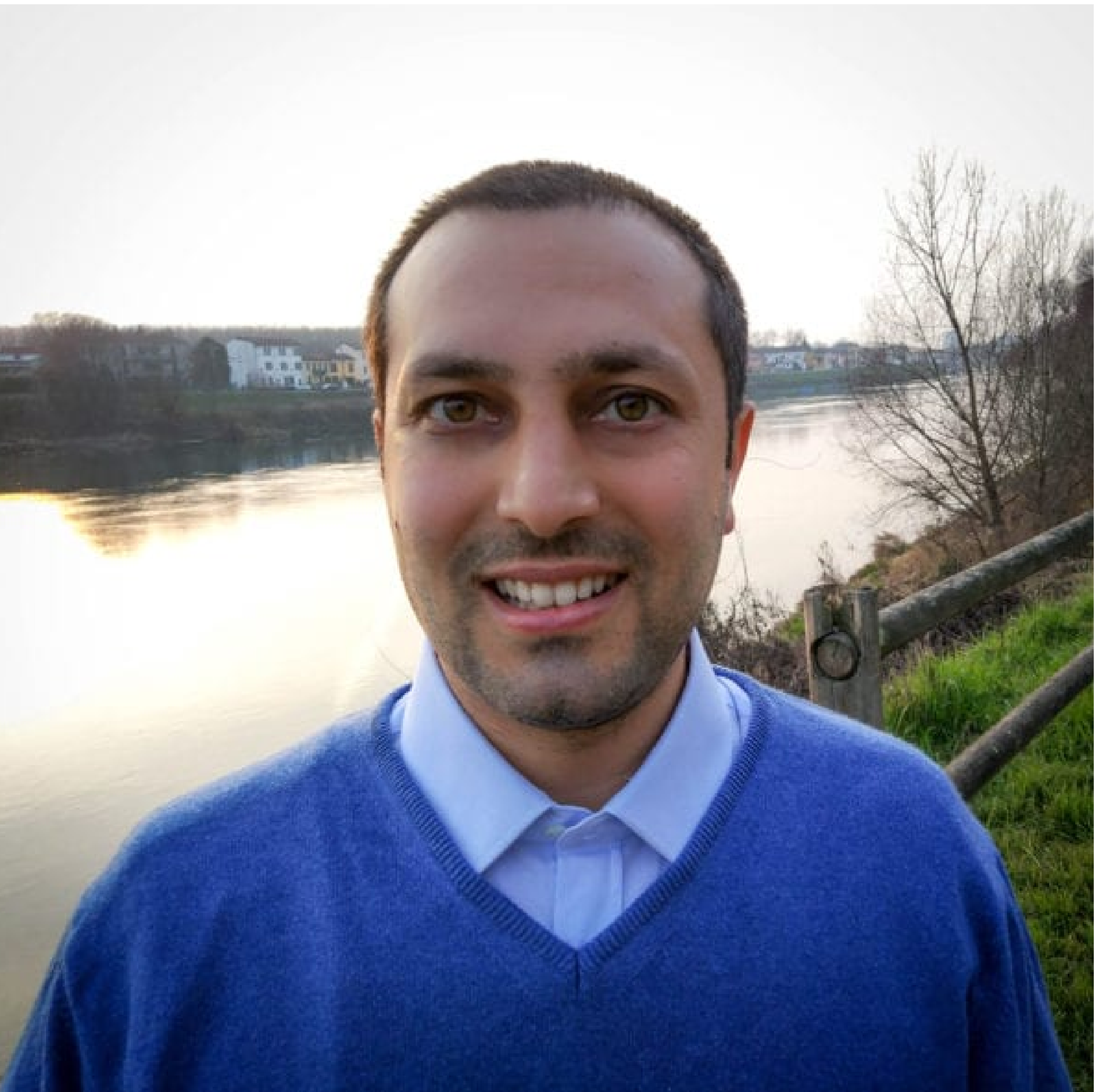}}]{Marco Castellano} received the Laurea degree from the Univ. of Pavia, Italy (2005) and in 2009 a Ph.D. in electrical engineering from Univ. of Pavia, Italy, in a joint research center supported by the Univ. of Pavia and STMicroelectronics. In 2008 he joined STMicroelectronics in Cornaredo (Italy) working in MEMS division as digital designer. His main fields of interest include complex gesture recognition algorithms implementation, FIFO, sensors, DSP and compensations design. Since 2016, he leads a team of digital experts working on co-design of controller and related software for custom low-power application design. He has authored several papers, conference contributions and patents on topics related to algorithms integration.

\end{IEEEbiography}

\begin{IEEEbiography}[{\includegraphics[width=1in,height=1.25in,clip,keepaspectratio]{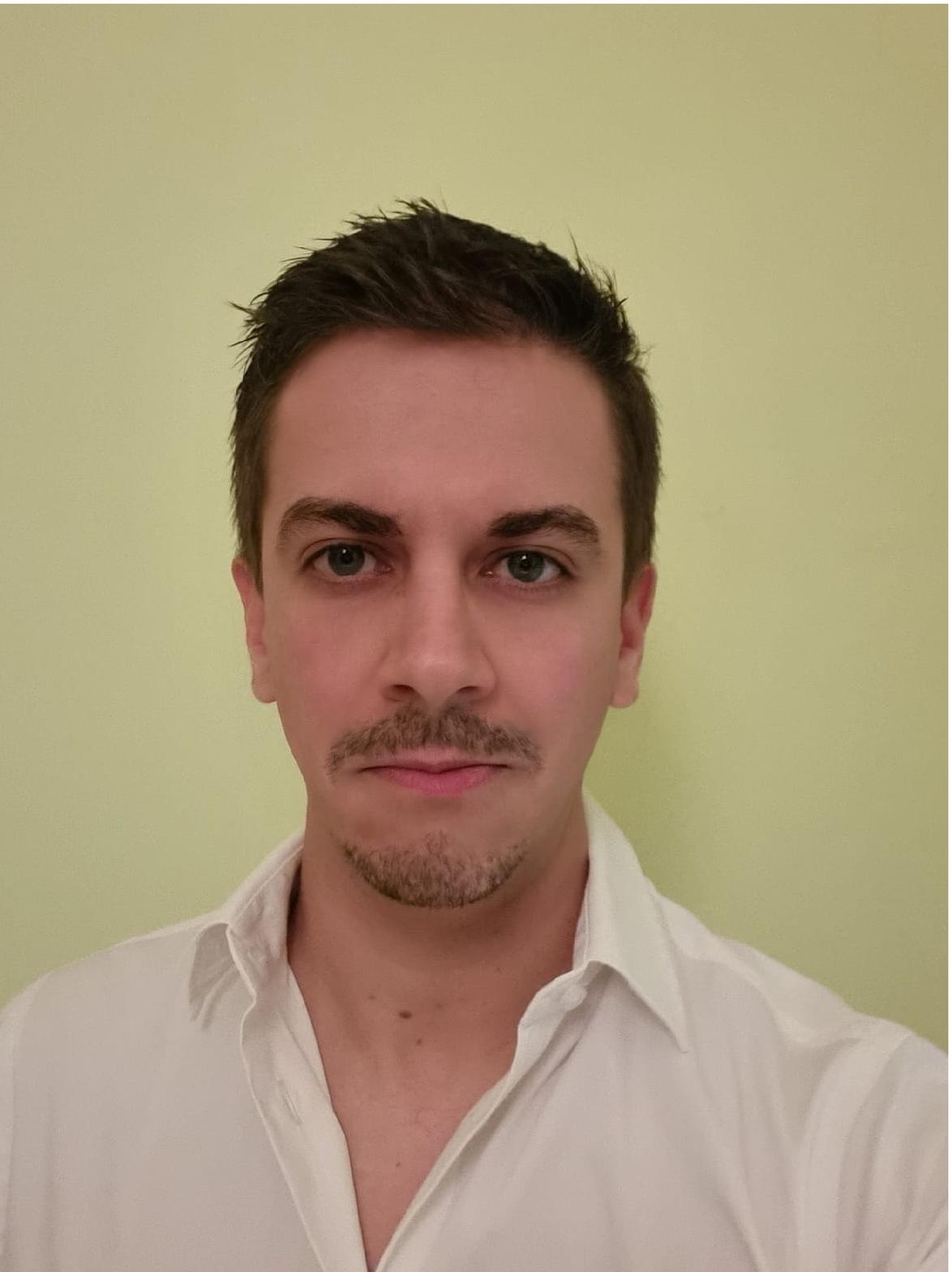}}]{Luca Gandolfi} received the Laurea degree from the Univ. of Pisa, Italy (2019). In 2019 he joined STMicroelectronics in Cornaredo (Italy) working in a R\&D digital design team for the Analog MEMS and Sensor Group. His research interest is in the codesign of firmware and hardware for complex algorithms in sensor systems.
\end{IEEEbiography}

\begin{IEEEbiography}[{\includegraphics[width=1in,height=1.25in,clip,keepaspectratio]{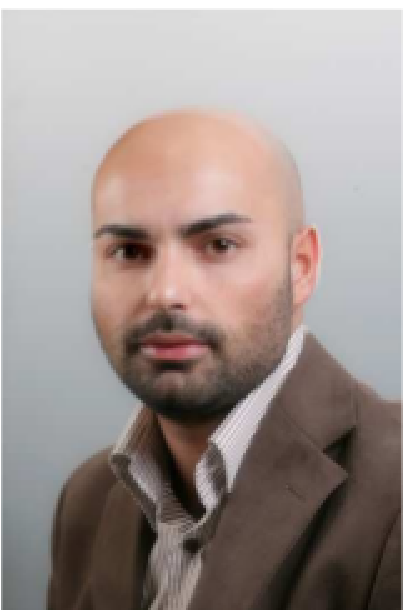}}] {Andrea Calimera} took the M.Sc. degree in Electronic Engineering and the Ph.D. degree in Computer Engineering from Politecnico di Torino. He is currently an Associate Professor of Computer Engineering at Politecnico di Torino. His research interests cover the areas of electronic design automation, with emphasis on optimization techniques for low-power and reliable integrated circuits, energy/quality management in embedded systems and portable applications, novel computing paradigms, and emerging technologies.
\end{IEEEbiography}

\begin{IEEEbiography}[{\includegraphics[width=1in,height=1.25in,clip,keepaspectratio]{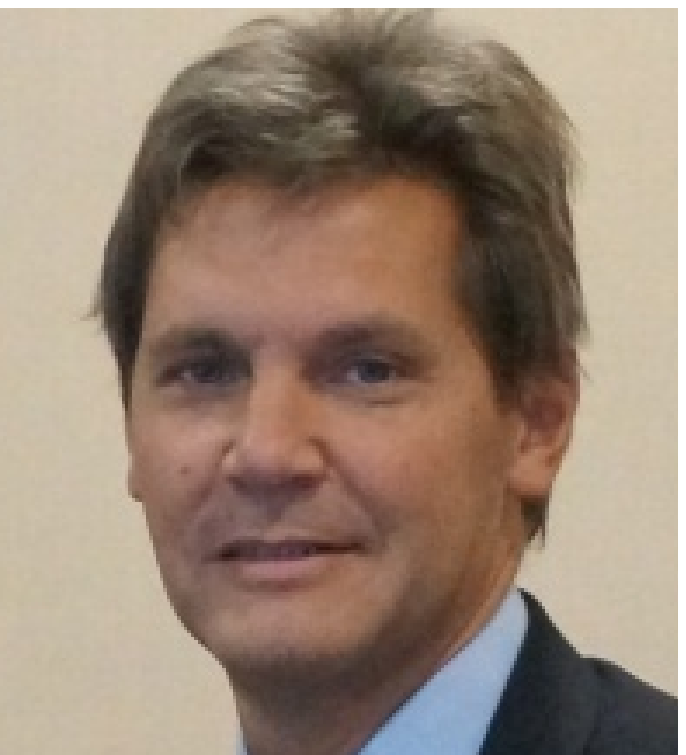}}]{Enrico Macii} is a Full Professor of Computer Engineering with the Politecnico di Torino, Torino, Italy. He holds a Laurea degree in electrical engineering from the Politecnico di Torino, a Laurea degree in computer science from the Università di Torino, Turin, and a PhD degree in computer engineering from the Politecnico di Torino. His research interests are in the design of digital electronic circuits and systems, with a particular emphasis on low-power consumption aspects energy efficiency, sustainable urban mobility, clean and intelligent manufacturing. He is a Fellow of the IEEE.
\end{IEEEbiography}

\begin{IEEEbiography}[{\includegraphics[width=1in,height=1.25in,clip,keepaspectratio]{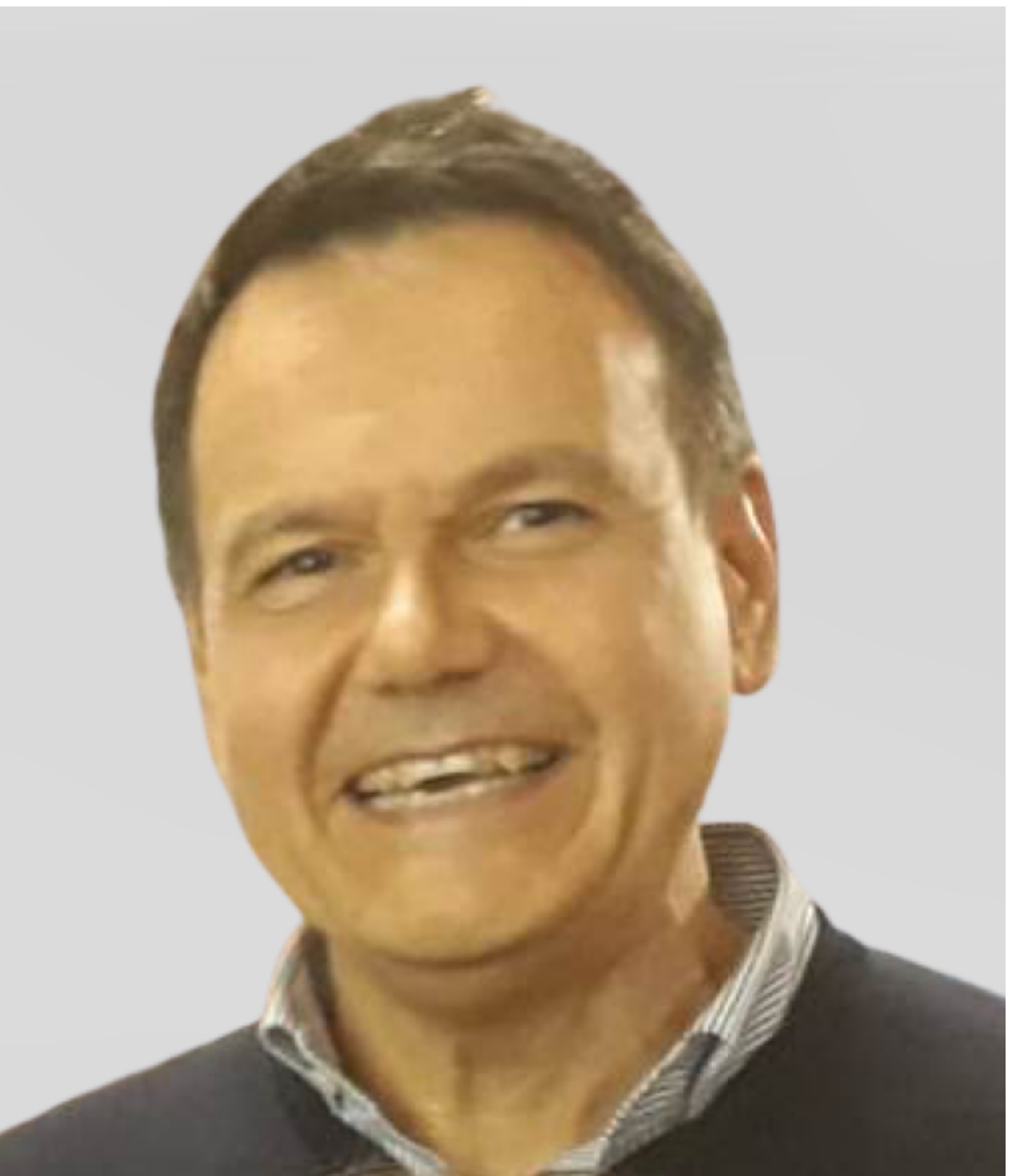}}]{Massimo Poncino} is a Full Professor of Computer Engineering with the Politecnico di Torino, Torino, Italy. His current research interests include various aspects of design automation of digital systems, with emphasis on the modeling and optimization of energy-efficient systems. He received a PhD in computer engineering and a Dr.Eng. in electrical engineering from Politecnico di Torino. He is a Fellow of the IEEE.
\end{IEEEbiography}

\begin{IEEEbiography}[{\includegraphics[width=1in,height=1.25in,clip,keepaspectratio]{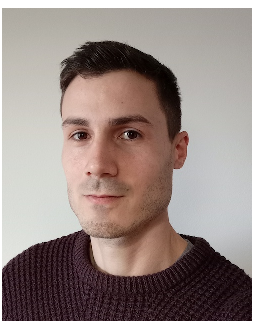}}]{Daniele Jahier Pagliari} received the M.Sc. and Ph.D. degrees in computer engineering from the Politecnico di Torino, Turin, Italy, in 2014 and 2018, respectively. He is currently an Assistant Professor with the Politecnico di Torino. His research interests are in the computer-aided design and optimization of digital circuits and systems, with a particular focus on energy-efficiency aspects and on emerging applications, such as machine learning at the edge.
\end{IEEEbiography}

\end{document}

%% file: sec/abstract.tex
Ultra-low-resolution Infrared (IR) array sensors offer a low-cost, energy-efficient, and privacy-preserving solution for people counting, with applications such as occupancy monitoring and visitor flow analysis in private and public spaces. Previous work has shown that Deep Learning (DL) can yield superior performance on this task. However, the literature was missing an extensive comparative analysis of various efficient DL architectures for IR array-based people counting, that considers not only their accuracy, but also the cost of deploying them on memory- and energy-constrained Internet of Things (IoT) edge nodes. Such analysis is key for system designers, since it helps them select the most appropriate DL model given the constraints of their target hardware. In this work, we address this need by comparing 6 different DL architectures on a novel dataset composed of IR images collected from a commercial 8x8 array, which we made openly available. With a wide architectural exploration of each model type, we obtain a rich set of Pareto-optimal solutions, spanning cross-validated balanced accuracy scores in the 55.70-82.70\% range. When deployed on a commercial Microcontroller (MCU) by STMicroelectronics, the STM32L4A6ZG, these models occupy 0.41-9.28kB of memory, and require 1.10-7.74ms per inference, while consuming 17.18-120.43 $\mu$J of energy. Our models are significantly more accurate than a previous deterministic method (up to +39.9\%), while being up to 3.53x faster and more energy efficient. So, our work serves also as a demonstration that DL can not only achieve higher accuracy, but also higher efficiency compared to classic algorithms for this type of task. Further, our models' accuracy is comparable to state-of-the-art DL solutions on similar resolution sensors, despite a much lower complexity. All our models enable continuous, real-time inference on a MCU-based IoT node, with years of autonomous operation without battery recharging.

%% file: sec/introduction.tex
\IEEEPARstart{D}{eep} learning (DL) has recently received attention in many Internet of Things (IoT) applications, ranging from embedded computer vision to time series forecasting, due to its remarkable predictive performance~\cite{chen2019deep,Jiang2019,Khan2020,Burrello2022a,Risso2022}. 
A direct execution of DL-based prediction tasks on \textit{extreme-edge} IoT nodes such as smart sensors can provide unique benefits compared with traditional cloud-based approaches, by eliminating the need of transmitting large amounts of raw data through a wireless network link~\cite{chen2019deep,Zhou2019,Shi2016}. Specifically, on-device execution makes the IoT node responsive even in bad or no-connectivity conditions, with a predictable latency. Moreover, the only information (optionally) transmitted to the cloud is the aggregated output of the DL model, e.g., a class label. This is beneficial for confidentiality, as it reduces the risk of accidental or malicious leakage of sensitive raw data (e.g., images, audio, video, etc)~\cite{chen2019deep,Zhou2019}.

However, DL algorithms originally designed for the cloud are energy-hungry and require high computational complexity, far beyond the capacity of memory- and energy-constrained IoT nodes, which are typically based on battery-operated and resource-limited Microcontrollers (MCUs). Bridging this gap in order to successfully deploy DL applications at the extreme edge requires a thorough selection of the employed models and of the corresponding hyper-parameters~\cite{Risso2022}.

Among the IoT applications that benefit from DL, people counting is increasingly popular due to its vast number of use cases in public safety, urban planning and commercial assistance~\cite{hou2010people}. Practical tasks range from monitoring the occupancy of indoor work spaces, museums and hospitals, to analysing the people flow statistics at the entrance of shops, supermarkets and other public places, to monitoring social distance violations or safety norms infringements especially in the context of the COVID-19 pandemic~\cite{tsou2020counting, xie2022privacy, perra2021monitoring}. 

There exist a wide range of technical solutions based on IoT for people counting, mainly split into two categories: instrumented and uninstrumented~\cite{raghavachari2015comparative}. The former approaches exploit the transceivers present in devices already owned by (or given to) users, such as smartphones, smartwatches, or tags~\cite{xi2014electronic}. However, these methods are heavily limited by voluntary participation and instrumental equipment, and are hard to apply in most real-world scenarios, especially in public places. On the other hand, uninstrumented solutions are free of the individuals' participation and rely on external sensors, such as proximity sensors, optical cameras, infrared arrays etc~\cite{hashimoto1997people, udrea2021new, raghavachari2015comparative, shami2018people}. 
Among these, infrared beam sensors and passive infrared sensors are inexpensive and simple to use, but rely on specific conditions such as object motion, and cannot easily distinguish multiple nearby people, which makes them often inaccurate~\cite{shetty2017detection}. As computer vision and video analysis techniques keep improving, vision-based people counting solutions are thus progressively replacing them.
Most current vision-based approaches use optical cameras, processing each frame with a Machine Learning (ML) algorithm to recognize and locate individuals~\cite{basalamah2019scale, nogueira2019retailnet, khan2019person}. While effective, they face severe privacy issues, since sensitive details of individuals such as facial information and body morphology are also recorded and processed.

In this scenario, \textbf{low-resolution infrared (IR) array} sensors offer a promising alternative, with advantages in terms of low energy consumption, low cost and privacy preservation. The latter is due to the fact that IR arrays only detect body temperatures, and given their low spatial resolutions (typically 8x8 or 16x16 thermal pixels), they can only capture the rough body shapes, hiding all privacy-sensitive details of individuals.
While other works have studied the combination of IR array sensors with DL models for people counting~\cite{bouazizi2022low, gomez2018thermal,metwaly2019edge,en14154542}, they: i) target higher resolution arrays, which simplifies the task but results in higher cost, higher energy consumption, and lower privacy and ii) consider a single type of DL model.

In this work, we perform the first detailed exploration and comparison of multiple DL model families for people counting based on a \textit{single, ultra-low-resolution (8x8)} IR array.  We focus on \textit{efficient} models, deployable on MCU-class platforms. 
The following is a summary of our main contributions:
\begin{itemize}
    \item We compare multiple efficient DL models for predicting the people count based on data from a single 8x8 IR array. For each type of model, we perform an extensive architecture exploration, obtaining a rich set of Pareto-optimal solutions in terms of performance and complexity.
    \item Analyzing the results of our exploration, we derive some interesting guidelines on the best type of model to prefer based on the target accuracy range and cost metric (model size or operations count). Overall, our models span a 55.70\%-82.70\% range in balanced accuracy, with parameters and operation counts varying in 0.4k-2.4k and 2.9k-20k respectively. The best balanced accuracy is up to 39.9\% higher than the one of a state-of-the-art deterministic algorithm~\cite{grideye}, and comparable with previous DL solutions on similar resolution data~\cite{bouazizi2022low}.
    \item We deploy some of the found models on a commercial MCU by STMicroelectronics, the STM32L4A6ZG, obtaining model size, inference latency, and inference energy values ranging in 0.41-9.28kB, 1.10-7.74ms and 17.18-120.43$\mu$J respectively. Our models are up to 3.53x faster and more energy efficient than~\cite{grideye}, while also being significantly more accurate. Furthermore, all of them allow real-time inference at 10 frames per second with very low energy consumption, which would permit years of continuous operation without battery recharging.
\end{itemize}

The rest of the paper is structured as follows: Section~\ref{sec:background} provides the background and overviews the related work on person counting applications based on IR sensors at edge. Section~\ref{sec:methods} presents a detailed description of the target dataset and of the various considered DL models, and describes the architecture exploration and deployment flow. Section~\ref{sec:results} reports the experimental results, and Section~\ref{sec:conclusions} concludes the paper.

%% file: sec/background.tex
\begin{table*}[t]
\centering
\caption{State-of-the-art people counting solutions based on infrared arrays\label{tab:literature}}
\resizebox{\textwidth}{!}{
\begin{threeparttable}
\renewcommand{\arraystretch}{1.2}
\begin{tabular}{|llllll|}
\hline
\textbf{Work}& \textbf{Sensor}& \textbf{Positioning} & \textbf{Dataset} & \textbf{Algorithm} & \textbf{Deployment Target}\\\hline\hline

Perra et al.~\cite{perra2021monitoring}  &  Grid EYE (8x8) & Door & Private  &  Deterministic & Z-Uno\\ \hline

Mohammadmoradi et al.~\cite{mohammadmoradi2017measuring} &  Grid EYE (8x8) & Door & Private & Deterministic & Raspberry Pi Zero\\ \hline

 Wang et al.~\cite{wang2021lightweight}  &    MLX90641 (12x16)  & Door & Private  &  Deterministic     & ESP8266 \\ \hline
 
Rabiee et al.~\cite{rabiee2021multi}  &  Grid EYE (8x8) & Ceiling & Private/Nagoya-OMRON Dataset~\cite{kawashima2017} & Deterministic & - \\\hline

Singh et al.~\cite{singh2019non}  &  MLX90621 (16x4) & Ceiling/Side Wall & Private & Deterministic & Arduino Uno \\\hline

Panasonic~\cite{grideye}  &    Grid EYE (8x8)  & Ceiling & LINAIGE~\cite{linaige} (*)  &  Deterministic  & STM32L4 (*)\\ \hline

Chidurala et al.~\cite{chidurala2021occupancy} &  \makecell[l]{ Grid EYE (8x8) \\  MLX90640 (32x24) \\ Lepton (80x60)} & Ceiling & Private &  \makecell[l]{Naive Bayes \\ KNN   \\  SVM  \\  RF } & Raspberry Pi 3 \\ \hline
 
Bouazizi et al.~\cite{bouazizi2022low}  &    MLX90640 (32x24)  & Ceiling & Private &  CNN   &  Raspberry Pi 3 \\ \hline

Gomez et al.~\cite{gomez2018thermal}  &    Lepton (80x60) & Wall & Private  &  CNN    & NXP LPC54102\\ \hline

Metwaly et al.~\cite{metwaly2019edge} &  MLX90640 (32x24) & Ceiling & Private & \makecell[l]{FNN \\ CNN \\  GRU} & STM32F4/F7 \\\hline

Kraft et al.~\cite{en14154542} &  MLX90640 (32x24) & Ceiling & Thermo Presence~\cite{en14154542} & CNN & Raspberry Pi 4 \\\hline

Xie et al.~\cite{xie2022privacy} &  Grid EYE (8x8) & Ceiling & LINAIGE~\cite{linaige} & CNN (2 variants) & STM32L4 \\\hline
Xie et al.~\cite{xieEnergyefficient2022} &  Grid EYE (8x8) & Ceiling & LINAIGE~\cite{linaige} & Wake-up Trigger + CNN & STM32L4 \\\hline\hline

\textbf{This Work} &  Grid EYE (8x8) & Ceiling & LINAIGE~\cite{linaige} & \makecell[l]{CNN (4 variants) \\ CNN-LSTM \\CNN-TCN} & STM32L4\\\hline

\end{tabular}
\begin{tablenotes}
\item[(*)] These entries refer to our deployment of the method described in~\cite{grideye}.
\end{tablenotes}
\end{threeparttable}
}
\end{table*}

\begin{figure}[t]
\centering
\includegraphics[width=.9\columnwidth]{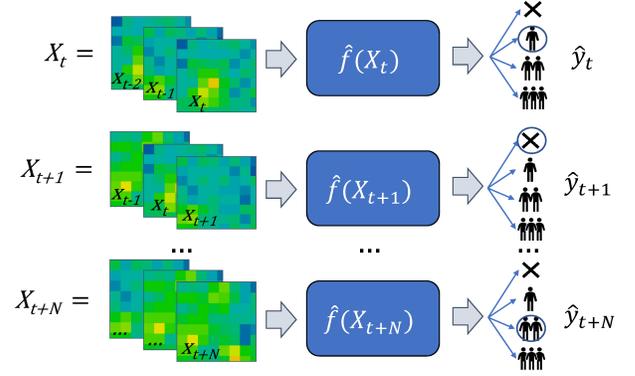}
\caption{People counting with IR array sensors: problem formulation. Depending on the work, the prediction function $\hat{f}(X)$ can be obtained either with a rule-based deterministic algorithm or learned from data using ML/DL, and the predicted person count $\hat{y}_{t}$ can be either a scalar or a class label.}
\label{fig:statement}
\end{figure}

People counting based on visual data is typically formulated as an object recognition problem~\cite{liu2005detecting}. Several sensor types have been utilized to implement both single- or multi-sensor systems for this task~\cite{perra2021monitoring, stec2019multi}. When considering this kind of sensor, the problem reduces to a classification or regression on image-like data, as shown in Fig.~\ref{fig:statement}. Namely, at time instant $t$, and calling $x_t$ the latest IR frame (i.e., ``image'') collected, the input to the recognition model is either a single frame $X_{t} = x_{t}$ or a \textit{window} of consecutive frames $X = \{x_{t-W+1}$,...,$x_{t}\}$, where $W$ is the window size ($W=3$ in the figure). The output is the predicted people count $\hat{y}_t = \hat{f}(X_{t})$, obtained either as a continuous scalar, then rounded to the nearest integer (regression formulation), or as a categorical value corresponding to one in a set of possible counts (classification formulation). The input/output relationship $\hat{f}(X)$ can be either obtained with a deterministic rule-based algorithm or learned from a training dataset using ML/DL approaches.

A summary of the most relevant literature works on people counting with multi-pixel IR arrays is reported in Table~\ref{tab:literature}. In particular, we report the sensor model and resolution, its position, the target dataset, the counting algorithms, and the IoT device considered in each work for deployment.
In detail, prior works leverage both deterministic algorithms~\cite{mohammadmoradi2017measuring, wang2021lightweight, perra2021monitoring, rabiee2021multi, singh2019non, grideye}, classic ML models~\cite{chidurala2021occupancy} or DL~\cite{bouazizi2022low, gomez2018thermal,metwaly2019edge, en14154542}. Among deterministic approaches, \cite{perra2021monitoring} implemented a novel real-time pattern recognition algorithm to process data sensed from doorway-mounted low-resolution IR array sensors to determine the number of people in a room. Similarly, \cite{mohammadmoradi2017measuring} also takes advantage of a doorway-mounted sensor, combined with a body extraction and localization algorithm, and background determination.
\cite{wang2021lightweight} proposed a similar lightweight deterministic solution based on a single array sensor positioned on a door, to monitor trajectories of objects entering and exiting a room, and estimate the indoor people count accordingly.
While interesting due to their use of a single, low-resolution sensor, these works solve a simplified and limited-scope version of the generic people counting problem.
In fact, they only permit the counting of people entering/exiting a room through a doorway. 

A more general deterministic method based on a ceiling-mounted sensor is described in~\cite{grideye}. 
This solution is based on the separation of moving thermal objects from the background by means of
smoothing, linear interpolation and hot area labeling and clustering. After that, threshold-based human detection is performed on each labeled thermal object to determine if it corresponds to a person or not. The reference background image is updated regularly to automatically filter stationary warm objects.

Furthermore, multi-sensor deterministic solutions have also been explored. Specifically, \cite{rabiee2021multi} proposed a people flow counting algorithm to monitor occupancy in smart buildings. To achieve this goal, multiple low-resolution sensors are deployed in connection points between different building areas, in order to count the number of people moving across adjacent zones.
The work of \cite{singh2019non}, instead, presents a framework to count people indoors based on two deterministic algorithms. Their method requires three 16x4 thermal sensors deployed at different locations, pointing to x, y, and z directions respectively.

Among classic ML works, \cite{chidurala2021occupancy} considers three ceiling-mounted IR arrays with different resolutions (8x8, 32x24, 80x60). It applies several preprocessing and feature extraction steps (active pixel and active frame detection, connected components analysis, statistical features), and then compares multiple classification algorithms for people counting. The considered algorithms are Naive Bayes, K-Nearest Neighbors (KNN), Support Vector Machines (SVM) and Random Forests (RFs). On a private dataset, they show that, for the lowest-resolution array, the best score is achieved with a RF.

Lastly, several DL-based solutions have been proposed. The authors of \cite{bouazizi2022low} use a Convolutional Neural Network (CNN) with 9 convolutional layers and 1 dense layer to process data from a ceiling-mounted, 32x24 pixels IR sensor to locate and count people indoors. Optionally, their proposed method allows the collection of lower-resolution samples (down to 8x6 pixels) to reduce sensor costs, thanks to the usage of a separate 8-layer CNN for frame upscaling.
\cite{gomez2018thermal} developed a head detection and people counting algorithm for wall-mounted sensors, based on a small-sized CNN model, and targeting a limited-memory low-power platform deployment, but focusing on a relatively high-resolution 80x60 pixels array.
\cite{metwaly2019edge} considered Feedforward Neural Networks (FNNs), CNNs and Gated Recurrent Units (GRU) for indoor occupancy estimation, based on ceiling-mounted 24x32 resolution IR arrays. 
The work of \cite{en14154542} also adopts a ceiling-mounted 24x32 resolution IR array, and leverages an encoder-decoder CNN architecture (a simplified version of U-Net) to reconstruct the position of people in the frame.

Most recently, in our previous work of~\cite{xie2022privacy}, we applied, to our knowledge for the first time, a DL model directly to the output of an ultra-low-resolution (8x8) array. However, that work considered a simplified version of the people counting problem, where the goal was simply to detect if the area covered by the sensor contained 2 or more people, in the context of social distance monitoring to combat the spread of COVID-19. The same task variant was tackled also in~\cite{xieEnergyefficient2022}, where an additional deterministic wake-up-trigger was used to avoid useless invocations to the CNN when no people are present in the frame, further reducing the energy consumption of the system.

All aforementioned data-driven (ML or DL) works suffer from important limitations: \cite{chidurala2021occupancy} and \cite{bouazizi2022low} only focus on deploying person counting on a high-end mobile Central Processing Unit (CPU), and they do not report detailed deployment results in terms of memory occupation of the models, inference latency, and energy consumption. \cite{gomez2018thermal} and \cite{metwaly2019edge} focus on relatively high-resolution arrays, which are more costly and power-consuming, besides possibly allowing the identification of users, thus reducing privacy. \cite{bouazizi2022low} supports low-resolution sensors only through an auxiliary CNN model for frame upscaling, which contributes to the total inference complexity. Furthermore, the excellent results obtained by many of these works~\cite{chidurala2021occupancy,gomez2018thermal,metwaly2019edge} are tainted by unfair data splitting, based on a random sampling at the level of individual frames or sliding windows. As explained in Sec.~\ref{sec:methods}, this unrealistically oversimplifies the task. The only work that performs a realistic data split at the session level is~\cite{bouazizi2022low}. Lastly, as mentioned, \cite{xie2022privacy} and \cite{xieEnergyefficient2022} focus on a simplified task variant.

In this work, we study for the first time the application of DL methods to a people counting problem based on the output on a \textit{single, ceiling-mounted, ultra-low-resolution IR array} (only 8x8 pixels). With an extensive architectural exploration of six families of efficient DL models, and many different hyper-parameters settings, we show that DL can not only provide significantly better counting performance compared to a deterministic algorithm, but also obtain benefits in terms of energy consumption, and latency.

%% file: sec/methods.tex
\subsection{Motivation}

The goal of this work is to perform a detailed exploration and comparison of various DL model families for people counting
based on a single, ultra-low-resolution (8x8) IR array. We focus on this setup due to its several practical advantages with respect to multi-sensor or higher-resolution alternatives, including better privacy preservation, lower overall system cost, and lower power consumption, especially for processing, as shown in our results of Sec.~\ref{sec:results}. In fact, intuitively, processing multiple and/or higher resolution images requires a higher number of operations, regardless of the specific algorithm employed, which is critical for ultra-low-power systems that need to operate for years on battery power.

As anticipated in Sec.~\ref{sec:introduction}, the main motivation for this study is that, to our knowledge, such an extensive comparison of DL models has not been performed before for this particular task. Therefore, we believe that it serves two related purposes: on the one hand, it provides a useful guidance for system designers that want to use this kind of sensor, for selecting an appropriate family of DL models based on the required accuracy and on the hardware memory, latency and energy constraints; on the other hand, it serves as a practical demonstration of the fact that DL can not only achieve higher accuracy, but also higher efficiency, compared to a classic algorithm~\cite{grideye}.

\subsection{Dataset}\label{sec:dataset}

There exists several public datasets containing IR array thermal images. However, most of them have been collected by relatively high-resolution sensors from 160 x 120 to 640 x 480, targeting applications such as pedestrian detection, and intelligent driving~\cite{olmeda2013pedestrian, flir, hwang2015multispectral, Rivadeneira_2020_CVPR_Workshops}. One public dataset containing low-resolution IR images is described in~\cite{baja-1j59-20}, and employs three wall-mounted sensors pointing in different directions, targeting human activity recognition tasks. Another low-resolution IR dataset containing 16x16 IR sensor arrays is described in~\cite{kawashima2017}, in this case for a ceiling-mounted sensor and specifically tailored for activity recognition. However, in this dataset at most one person appears in the frame, which deviates from the original people counting purpose.
The dataset in~\cite{en14154542} is instead dedicated to people counting applications, with up to 5 people in one frame. Each frame is annotated with the people's locations, which can be simply converted into counts, but the dataset is collected with a relatively high-resolution array (24x32 pixels).
None of these datasets are suitable for experimenting on low-cost, energy-efficient people counting on ultra-low-resolution IR arrays. Indeed, as shown in Table~\ref{tab:literature}, most literature on this task uses privately-collected data.

\begin{figure}[t]
\centering
\includegraphics[width=0.8\columnwidth]{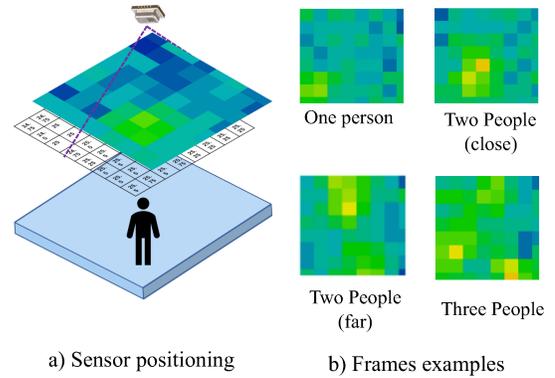}
\caption{Sensor mounting and example of the IR frames.}\label{fig:sensor}
\end{figure}

Given this scenario, we collected and made openly available a new dataset called LINAIGE (Low-resolution INfrared-array data for AI on the edGE)~\cite{linaige}. LINAIGE targets specifically people counting and presence detection tasks in indoor environments, and its first version was described in our previous work of~\cite{xie2022privacy}. The dataset includes IR samples collected with a Panasonic Grid-EYE (AMG8833) sensor~\cite{grideye} outputting a 8 x 8 array, at 10 Frames Per Second (FPS). Each frame is associated with the corresponding people count label.
During data collection, the sensor was ceiling-mounted as shown in Fig.~\ref{fig:sensor}a, and positioned in different indoor environments such as offices, laboratories and corridors, using a lens with a view angle of 60°. 
Volunteers passed in the view range of the sensor by walking, standing, running, etc, during a number of data collection sessions. Some examples of collected frames and corresponding people counts are shown in Fig.~\ref{fig:sensor}b. As detailed in~\cite{xie2022privacy}, depending on the sensor height in different environments, the maximum distance between in-frame people varies in [1.53:2.04] m and the counting area is up to $\approx$ 2 m\textsuperscript{2}. People counting on larger areas can be simply achieved by combining the outputs of multiple sensors, appropriately positioned.
With respect to the original dataset described in~\cite{xie2022privacy}, this work is based on a new version with improved data quality. Namely, we removed the very rare frames with $>3$ people (0.66\% of the total), which were present only in one session, complicating the training and cross-validation of ML/DL models. Further, we also removed the shortest session (session 4 in~\cite{xie2022privacy}) which contained only 196 frames, i.e. around 20s worth of data, and unrealistically altered the recognition performance metrics. After these changes, the new dataset contains 25110 samples, split into 5 sessions. Each session is associated with a timestamp, environment name and room temperature.

\begin{table*}{}
    \centering
     \caption{Dataset statistics and Cross Validation strategy.}\label{tab:cv}
    \scriptsize
    \begin{tabular}{|c|c|c|c|c|c|c|c|c|c|c|c|c|c|}
    \hline
    \multicolumn{6}{|c}{\textbf{Train Fold}} &  \multicolumn{6}{|c|}{\textbf{Test Fold}}  \\ \hline
    \multirow{2}*{Session} & \multirow{2}*{Sample N.} & \multicolumn{4}{c|}{People Counts Statistics [\%]} & \multirow{2}*{Session} & \multirow{2}*{Sample N.} &  \multicolumn{4}{c|}{People Counts Statistics [\%]} \\ \cline{3-6} \cline{9-12}
    ~ &  ~ & 0 & 1 & 2 & 3 & ~ & ~ & 0 & 1& 2& 3 \\ \hline
    1, 3, 4, 5 & 23529  & 26.07 & 43.49 & 23.61 & 6.83  & 2 & 1581 & 14.86 & 30.68 & 54.46  & 0 \\ \hline
    1, 2, 4, 5  & 23591 & 22.37 & 44.03 & 26.84 & 6.77 &3 & 1519 & 71.89 & 21.72 & 5.66 & 0.72 \\ \hline
    1, 2, 3, 5  & 22908 & 25.3 & 41.85 & 26.17 & 6.67 &4 & 2202 & 26.02 & 51.27 & 19.16 & 3.54\\ \hline
    1, 2, 3, 4  & 23260 & 24.69 & 43.02 & 26.08 & 6.20 &5 & 1850 & 33.78 & 38.38 & 18.92 & 8.92\\ \hline
     \end{tabular}
\end{table*}

IR frames have been labelled using a semi-automatic method: a data collection system based on a single-board computer named Raspberry Pi 3B has been set up, including both the IR sensor and an optical camera, pointing in the same direction and collecting synchronized frames. Optical frames have been then processed with a pre-trained object detection model (Mask R-CNN~\cite{He2017}) to automatically count the number of people in them, and associate the same count to the corresponding IR frame. The results have been double-checked by a human labeller to correct CNN mispredictions. 
Further, the human labeller also associated each frame with a binary \textit{confidence} measure, which can be used to exclude frames for which it was difficult to assess the exact people count due to the imperfect alignment of the viewing angles between the IR sensor and the optical camera. More details on the labelling are found in~\cite{xie2022privacy}.

In all experiments of this work, we excluded ``hard-to-label'' frames from training and testing, both for our method and for state-of-the-art comparisons.
Moreover, in contrast to~\cite{xie2022privacy} where a simple per-session train/test split was used, here we adopt a per-session Cross Validation (CV) approach, to make our model evaluation independent from the characteristics of a specific test session. The cross validation strategy is shown in Table \ref{tab:cv}. Given that Session 1 is significantly larger than all others (17958 frames versus a maximum of 2202 for other sessions, and 71\% of the total data), we always kept it in the training set. Sessions 2, 3, 4 and 5 have been rotated as the test set in different iterations, with all other data in the training set, yielding 4 CV folds. 
This \textit{leave-one-session-out} CV strategy ensures the fairness of model evaluation, by making sure that test frames correspond to a different environment, date-time, and room temperature setting compared to training frames. This is close to a realistic scenario, in which the system is likely to be tested in a different environment from where it was trained. In contrast, a purely random per-frame split would cause a leakage of information between training and testing, oversimplifying the problem.

\subsection{Model Architectures}

We considered six families of DL models to predict the people count in IR frames, exploring some of the key hyper-parameters of each. A graphic representation of all considered models is shown in Fig.~\ref{fig:cnn}.

\subsubsection{Single-frame CNN}\label{sec:single_frame_cnn}

\begin{figure*}[ht]
\centering
\subfigure[Single-frame CNN]{\includegraphics[width=0.48\textwidth]{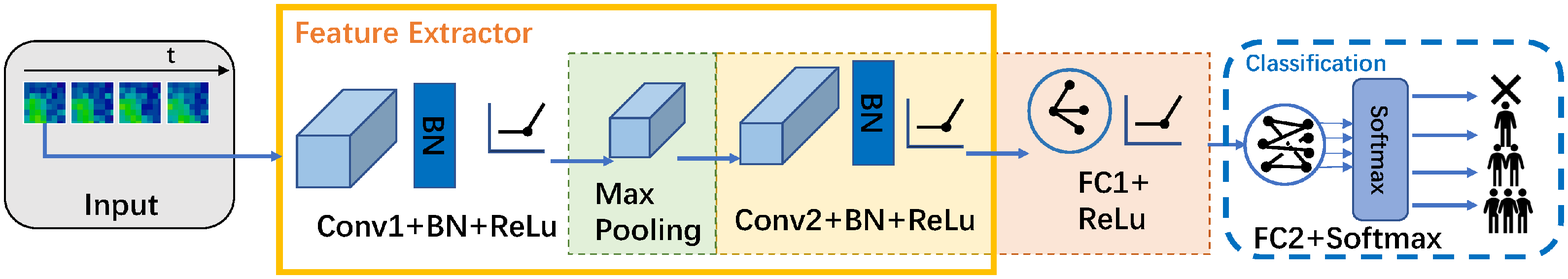}}
\subfigure[Multi-channel CNN]{\includegraphics[width=0.48\textwidth]{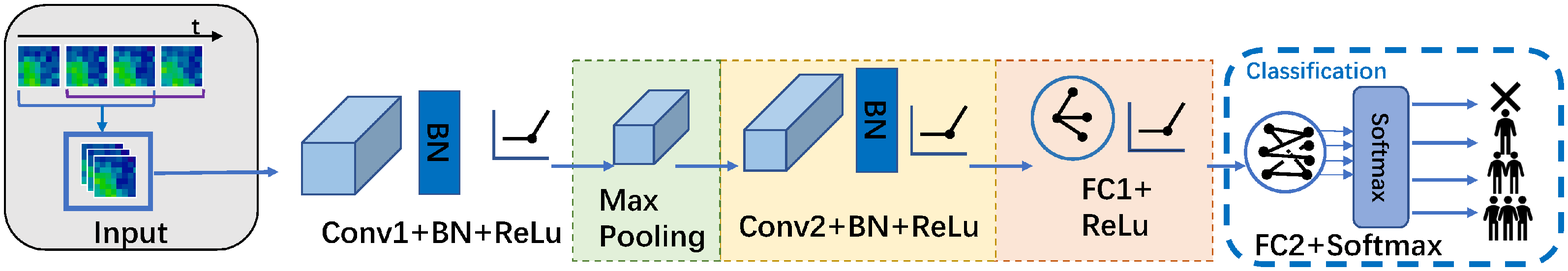}}
\subfigure[Majority Voting CNN]{\includegraphics[width=.48\textwidth]{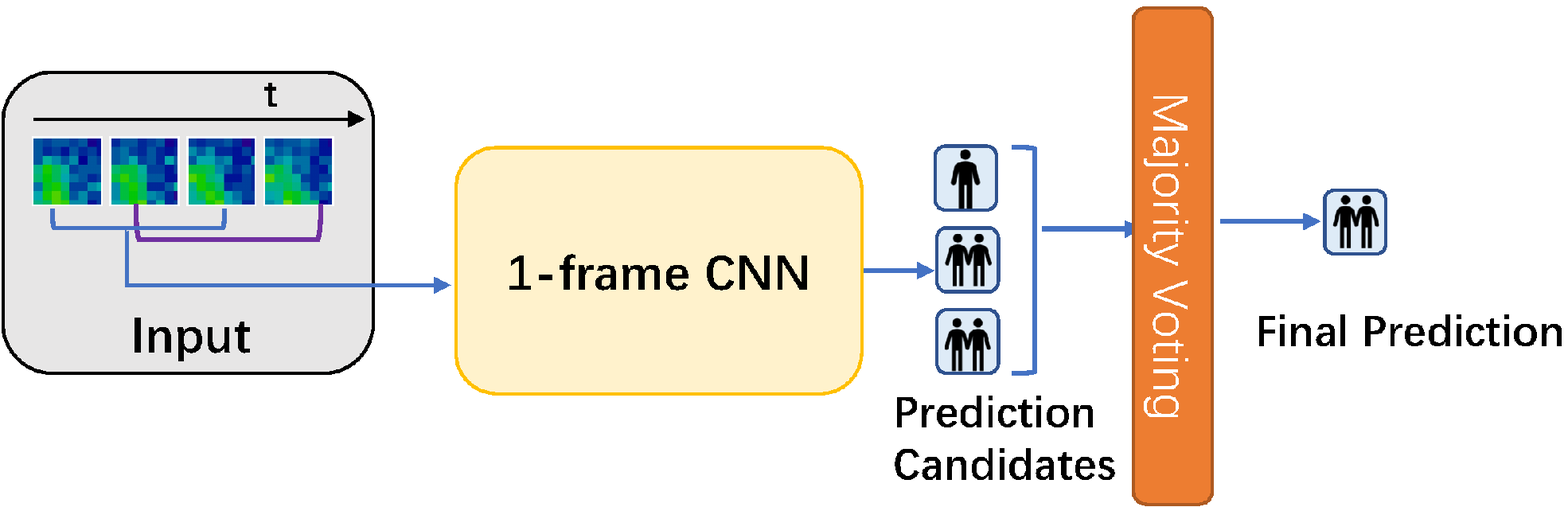}}
\subfigure[Concatenated CNN]{
\includegraphics[width=.48\textwidth]{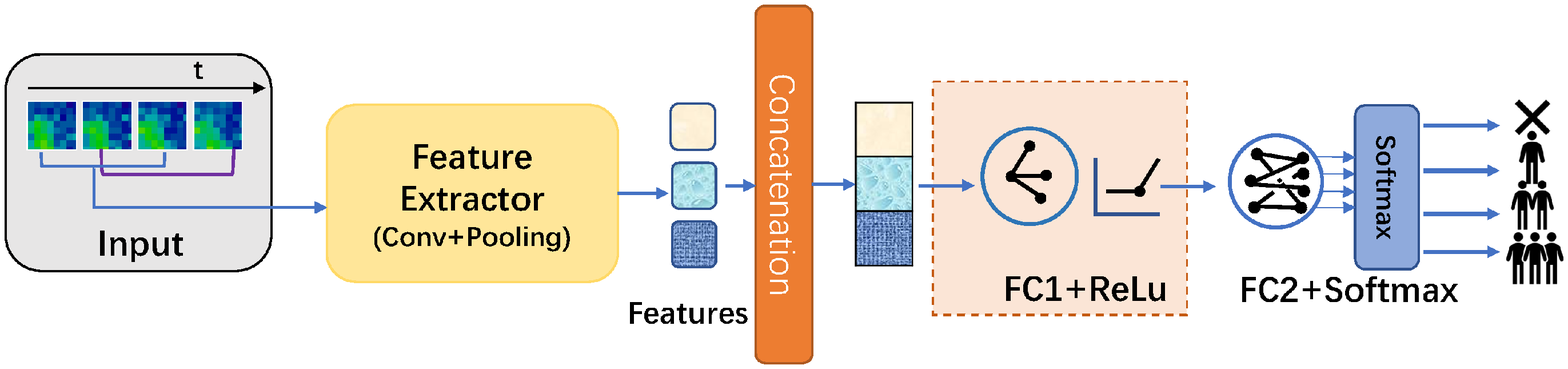}}
\subfigure[CNN-LSTM]{
\includegraphics[width=.48\textwidth]{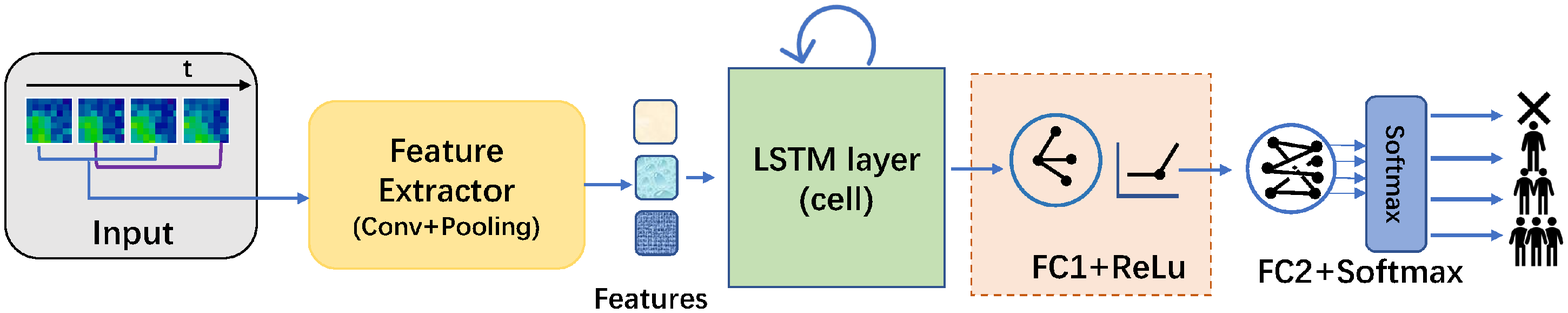}}
\subfigure[CNN-TCN]{
\includegraphics[width=.48\textwidth]{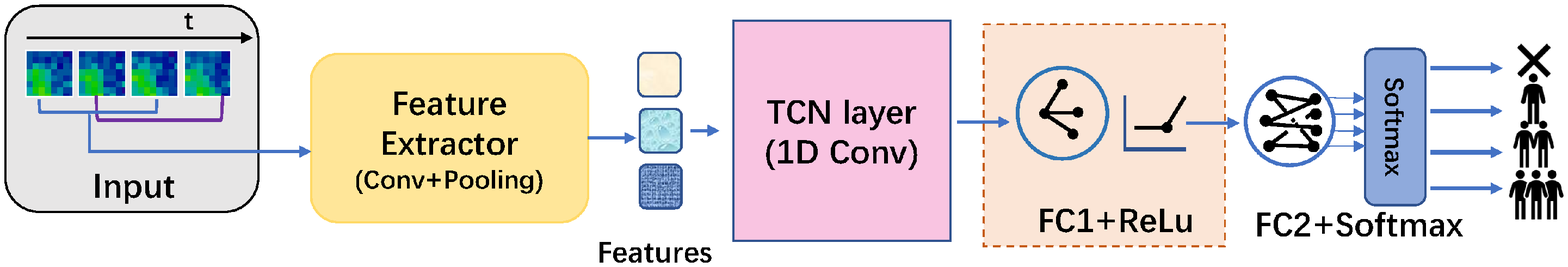}}
\caption{Model Architectures considered in this work}
\label{fig:cnn}
\end{figure*}

The first considered architecture is a simple CNN, which is known to be effective in many image-based pattern recognition tasks.
The general template of the considered CNNs is shown in Figure~\ref{fig:cnn}a; it includes up to 2 Convolutional (Conv) layers with Rectified Linear Unit (ReLU) activation, 1 optional Max Pooling layer and up to 2 Fully Connected (FC) layers. The first FC layer has 64 hidden units and a ReLU activation, while the output layer has a number of neurons equal to the possible count ``classes'' (from 0 to 3 people, corresponding to 4 output neurons, in our experiments).
Furthermore, compared to our previous work of~\cite{xie2022privacy}, which focused on a simpler social distancing problem, we added Batch Normalization (BN) layers after each Conv layer to improve the classification performance.
Utilizing this template as a starting point, a vast architecture exploration was performed, by eliminating/retaining layers which are enclosed in dashed boxes in Fig.~\ref{fig:cnn}a. Namely, we considered architectures with:

\begin{itemize}
    \item 1 or 2 Conv layers, each followed by BN; 
    \item 1 or 2 FC layers;
    \item 0 or 1 Max Pooling layers;
\end{itemize}
Besides varying the number of layers, we also explored the number of feature maps (i.e., \textit{channels}) in each Conv layer, considering values in \{8, 16, 32, 64\}.
Conv. and Pooling kernel sizes are fixed at 3x3 and 2x2 respectively.
The input processed by this CNN model is a \textit{single} IR array frame $X_t = x_t$, with a tensor shape (8, 8, 1). In total, we evaluate 48 different Single-frame CNN variants.

\subsubsection{Multi-channel CNN}
\begin{figure}[t]
\centering
\includegraphics[width=\columnwidth]{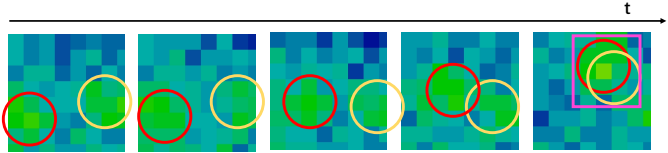}
\caption{Example of the IR frames sequence corresponding to 2 people moving close to each other.}\label{fig:win_example}
\end{figure}

While the previous model considers a single IR array frame as input, all other models try to exploit the temporal information enclosed in a \textit{sequence} of consecutive frames to improve the people counting accuracy. The rationale is that considering a sliding window of IR frames as input can reveal information on people movement, which in turn can improve the prediction accuracy in complex cases. For instance, Fig.~\ref{fig:win_example} shows that a single hot area (highlighted by a purple box in the last frame) can be correctly associated with two people close to each other, rather than with a single person, by observing the movement of the two people (red and orange cycles) in preceding frames.   

The first and simplest mechanism that we considered to process multiple IR frames consists in feeding them to a CNN as \textit{different input channels}. Specifically, calling $W$ the length of the sliding window, a tensor with shape (8, 8, W) is formed by stacking IR frames $X_t = \{x_{t-W+1}$,...,$x_{t}\}$ along the channels axis. The tensor is then associated with the people count label of the \textit{last} frame $y_{t}$ for training and testing.
These inputs and outputs are also used for all the other multi-frame architectures described in the following. The template of the Multi-channel CNN model is shown in Fig.~\ref{fig:cnn}b. 
We explored the same hyper-parameters settings considered for Single-frame CNNs in terms of the number of layers, and the number of Conv channels.
In addition, we also varied the window size $W$ in \{3, 5, 7, 9\}. This exploration is interesting because, intuitively, with a too-short window the advantages of accessing past frames are limited, whereas a too-long window will provide useless information (too far in the past), while increasing the time and memory complexity of the first Conv layer.

\subsubsection{Majority Voting CNN}

Majority voting is a simple yet effective ensemble learning approach that takes advantage of multiple classification results to generate final predictions with lower variance~\cite{lam1997application}.  In recent years, several literature works have applied this technique, using either different classifiers~\cite{lee1993handprinted}, multiple instances of the same model trained differently~\cite{amin2020cnns} or a single trained model fed with different inputs~\cite{Yazdizadeh2020}.
In our work, we follow the latter approach, applying majority voting (i.e., \textit{mode inference}) to the $W$ predictions obtained by executing a Single-frame CNN on each frame of the sliding window. 
A high-level scheme of this solution is shown in Fig.~\ref{fig:cnn}c. The clear advantage of this technique, from the point of view of edge inference, is that it requires approximately the \textit{same memory} as a single-frame CNN, while possibly improving the prediction accuracy by filtering-out occasional mispredictions. The latency/energy cost for inference, instead, is roughly $W$ times higher than that of a single-frame model.
We consider again $W$ values in \{3, 5, 7, 9\}. Note that majority voting requires an odd window size; thus, for fairness of comparison, all other multi-frame architectures have been tested only with odd $W$ values.

To make our architectural exploration tractable, we consider the majority-voting models obtained using \textit{Pareto-optimal Single-frame CNNs} as individual predictors. More specifically, we apply majority voting on top of all single-frame CNNs found in the Pareto front in terms of people counting accuracy versus model size or versus number of operations.

\subsubsection{Concatenated CNNs}
While the main advantage of majority voting is that it does not require extra trainable parameters, its main drawback is that it cannot assign different \textit{importance} to the various IR frames in the sliding window. Intuitively, more recent frames should be given more importance to determine the people count, especially with large $W$. 
Although this could be approached by a \textit{weighted voting} mechanism, such solution requires a difficult hand-tuning of the weights assigned to each frame.
Thus, our next considered DL model exploits a feature concatenation approach to overcome this limitation~\cite{demir2020new, wu2019concatenate}. 
Specifically, as illustrated in Fig.~\ref{fig:cnn}d, each frame of the sliding window is individually fed into a feature extractor module to extract time-independent features. Then, all outputs are flattened and concatenated into a unique feature vector, and further processed by two FC layers to generate the final prediction. In this way, the training process can automatically assign appropriate weights to different frames' features.
We consider the Conv and Pooling layer configurations (i.e., the part highlighted by an orange box in Fig.~\ref{fig:cnn}a) found in each Pareto-Optimal Single-frame model as possible feature extractors for concatenated CNNs. 
Furthermore, besides exploring the usual 4 values of $W$, we also vary the number of neurons in the first FC layer in \{8, 16, 32, 64\}. Altogether, given $N$ Pareto-optimal feature extractors, we evaluate a total of 4*4*N Concatenated CNNs.

\subsubsection{CNN-LSTM}
The next multi-frame model explicitly considers the time dependency between frames, replacing the simple feature concatenation with a Long-Short Term Memory (LSTM) cell. Several works have considered CNN-LSTM models to combine spatial and temporal information~\cite{kim2019predicting, zhao2019speech, sciannameoDeep2022}. The closest work to ours is \cite{kawashima2017}, which applied a CNN-LSTM for human activity recognition based on a 16x16 IR array. These works have demonstrated the remarkable performance achieved by CNN-LSTMs. However, LSTM cells are less hardware-friendly than CNNs~\cite{leaTemporal2016} (see Sec.~\ref{sec:cnn_tcn}). Therefore, it is interesting to compare this model with other architectures, considering the trade-off between complexity and performance.

Our CNN-LSTM template is shown in Fig.~\ref{fig:cnn}e. The $W$ feature extractor outputs are flattened and fed to the LSTM cell sequentially. One or two FC layers are then connected to the last hidden state produced by the LSTM to generate the output prediction.
We apply the same feature extractors selection strategy illustrated above for Concatendated CNNs. Moreover, we vary $W$ as before, and we also explore the number of hidden units in the LSTM cell, with values in \{8, 16, 32, 64\}. Again, given $N$ Pareto-optimal feature extractors, a total of 4*4*N CNN-LSTM architectures are evaluated.

\subsubsection{CNN-TCN}\label{sec:cnn_tcn}
The last type of model considered is based on Temporal Convolutional Networks (TCN)~\cite{leaTemporal2016} which have recently emerged as a more hardware-friendly alternative to LSTMs, and have been applied to several edge-relevant tasks~\cite{Burrello2021c,Risso2022}. TCNs are simply 1D CNNs, with the peculiarity of using \textit{causal} convolution, which is appropriate for time-series processing. Compared to LSTMs, these networks exhibit more data reuse and are more resilient to integer quantization, both of which are advantageous for edge deployment~\cite{leaTemporal2016}. Therefore, our last architectural template is built by combining the outputs of the usual 2D CNN feature extractors applied to single IR frames with a single TCN layer, as shown in Fig.~\ref{fig:cnn}f. The TCN output is then flattened and fed to 1 or 2 FC layers to generate a prediction.  We fix the 1D Conv kernel size at 3x1, and the dilation at 1.
Besides varying $W$ as in previous models, we explore the number of output channels of the TCN layer, considering values in \{8, 16, 32, 64\}. Therefore, with $N$ feature extractors, also in this case we explore 4*4*N architectures.

\subsection{Training and Deployment Flow}\label{sec:deployment}

All models are trained with the leave-one-session-out CV strategy described in Sec.~\ref{sec:dataset}. At first, we perform a standard floating point model training with Keras/TensorFlow 2.0~\cite{tensorflow}, for a maximum of 500 epochs per fold. 
We optimize a categorical cross-entropy loss function using the ADAM optimizer, with an initial learning rate of $10^{-3}$.
A learning rate reduction of factor 0.3 is applied when the training loss is stagnating, with a patience of 5 epochs.
Early-stopping is applied after 10 non-improving epochs.
Given the strong class imbalance of the LINAIGE dataset (see Table~\ref{tab:cv}), we apply class-dependent weights to the loss during training, which are computed as the inverse of the class frequencies. 

After this initial floating point training, we quantize the parameters, inputs, outputs, and intermediate activations of the resulting models to 8-bit integers, using the TensorFlow Model Optimization (TFMOT) API. This step is important to further reduce the memory occupation, latency, and energy consumption of the models, when deployed on constrained MCU-based IoT nodes~\cite{Daghero2021a}. We then apply quantization-aware training (QAT)~\cite{Jacob2018} to recover the accuracy drop due to quantization as much as possible. We use the same training protocol described above, with the only two differences that the initial learning rate is set to $5\times 10^{-4}$ and the learning rate scheduling and early stopping patience values are set to 10 and 20 epochs respectively.
Note that the QAT of LSTM cells is not supported by the TFMOT API yet. Therefore, the CNN-LSTM models are directly deployed in floating point to the MCU. This turns out to be a major practical limitation of CNN-LSTMs.

The trained and quantized models are then converted into TensorFlow Lite (TFLite) format~\cite{tensorflow}. Lastly, we utilize the X-CUBE-AI toolchain 7.2.0~\cite{cubeai} to convert the TFLite files into optimized C language implementations for our deployment target, i.e., the ultra-low-power STM32L4A6ZG MCU by ST Microelectronics, which is based on a 32-bit Cortex-M4 core~\cite{mcu}. The latency and energy results refer to the MCU running at 80MHz, with a supply voltage of 1.8V.

%% file: sec/results.tex
\begin{figure*}[t]
\centering
\includegraphics[width=0.9\linewidth]{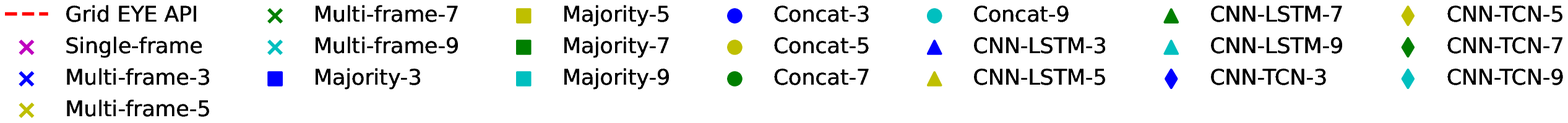}
\includegraphics[width=\linewidth]{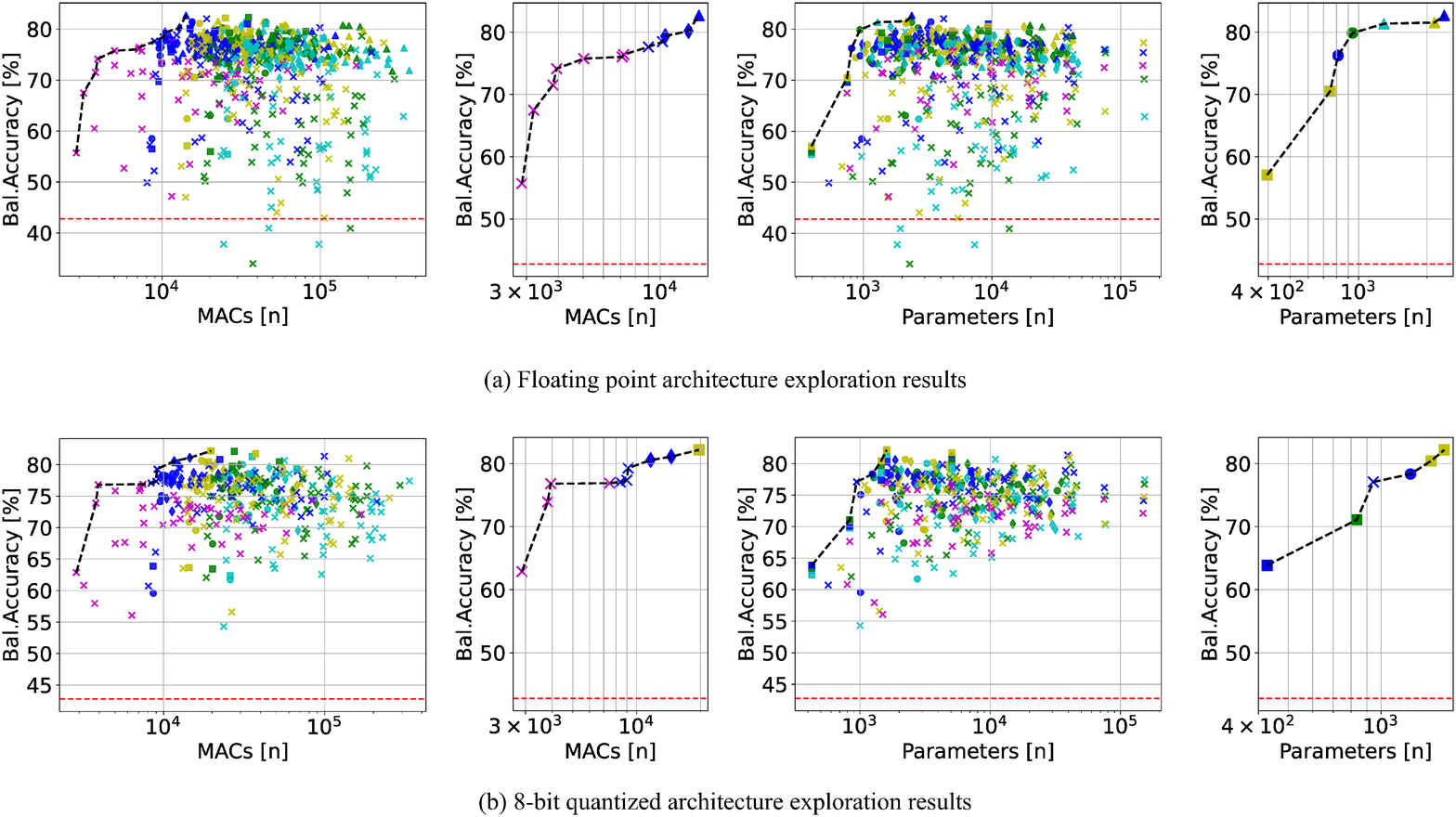}
\caption{Results in terms of balanced accuracy versus number of parameters and number of MAC operations for all considered models. All models (left), and isolated Pareto fronts (right).}\label{fig:all}
\end{figure*}

\subsection{Setup}

To evaluate the performance of our models, we mainly consider the Balanced Accuracy (Bal. Acc.) metric, i.e., the average of recall on each class. Compared to the standard accuracy (Acc.), i.e., the fraction of correct predictions, which we also report for completeness, the Bal. Acc. is more suitable for class-unbalanced datasets. Moreover, we also measure the F1-Score (F1), defined as the harmonic mean of precision and recall. Since ours is a multiclass problem, we compute the weighted average of the F1 on each class. Lastly, we also report the Mean Absolute Error (MAE) and the Mean Squared Error (MSE) between ground truth and predicted people counts.
We consider MAE and MSE although our task is a classification, because they allow taking into account the \textit{significance} of errors: e.g., for a frame with a ground truth people count of 3, a model that outputs 2 makes a ``smaller'' error compared to one that outputs 1.
All metrics are reported as the mean $\pm$ standard deviation over the 4 CV folds, where each fold is weighted by the number of its test samples over the total test samples.

To estimate the hardware-independent computational complexity of each model, we consider the \textit{number of parameters} as a proxy for model size, and the \textit{number of Multiply-and-Accumulate (MAC)} operations, i.e., the dominant operations in DL inference, as a proxy for energy and latency. 
We then deploy on the target MCU a selection of Pareto-optimal models in the Bal. Acc. versus parameters and MACs planes.
For deployed models, we derive the total memory occupation, as well as the total clock cycles, energy consumption and latency per prediction, from measurements on the real hardware, i.e, the STM32L4A6ZG MCU by ST Microelectronics. Concerning memory occupation, both the model size and the total Flash usage are measured. In particular, model size is obtained from the X-CUBE-AI toolchain when generating C code for our DL models, while Flash usage is evaluated using the STM32CubeIDE~\cite{stcubeide} during deployment. The CPU cycles per inference, which determine the total latency, are also measured using STM32CubeIDE, and in turn used to compute the energy consumption based on the average active power of the MCU from the datasheet. We consider the STM32L4A6ZG MCU working at 80MHz clock frequency, with 1.8V supply voltage~\cite{mcu}. The latency and energy consumption estimates for each architecture have been obtained running the models and the baselines 1000 times, and reported as the mean $\pm$ standard deviation, as shown in Table \ref{tab:deploy}. 

Our main baseline for comparison is~\cite{grideye}, i.e., the only publicly available people counting solution based on a ceiling-mounted IR array with the same resolution sensor as ours. We compiled and executed the code of~\cite{grideye}, written in C language, on our target MCU, using the same compilation flags of our models, and we tested it on the LINAIGE dataset. Furthermore, we also compare with~\cite{chidurala2021occupancy,bouazizi2022low,gomez2018thermal,en14154542,metwaly2019edge}, although only qualitatively, since those works target different datasets and hardware platforms.

\subsection{Architecture Exploration}\label{sec:exploration}

Figure~\ref{fig:all} shows the results of our architecture exploration. In particular, the top (bottom) graphs show the results before (after) 8-bit quantization. For each data precision and target cost metric (MACs or parameters), we report both the entire set of considered models (left), and a ``zoom'' on the Pareto frontier (right), highlighted by a black dashed line. The people counting performance is reported in terms of the average Bal. Acc. over the CV test folds. Each marker shape refers to one model type, whereas colors correspond to different sliding window sizes $W$. Note that LSTM-based models are not present in Fig.\ref{fig:all}b because their quantization is not supported by TFMOT. The performance of our comparison baseline of~\cite{grideye} is shown by a horizontal red line.

The complete graphs show the breadth of our architectural exploration, which includes models that span more than two orders of magnitude in terms of MACs (2.9k-364k) and parameters (0.4k-153k). When considering only Pareto-optimal models, the MACs range is 2.9k-14k for float models and 2.9k-20k for quantized models, while parameters vary in 0.4k-2.4k and 0.4k-1.6k respectively.
The Bal. Acc. spanned by these models ranges in 55.70-82.70\% for float models and 62.88-82.17\% for quantized ones.

All Pareto-optimal models outperform the deterministic approach of~\cite{grideye}, showing the benefit of data-driven methods for this task. Moreover, all 6 considered model families are present in at least one Pareto front, demonstrating that focusing on a single architectural template would be sub-optimal.
Single-frame CNNs are only achieving optimal trade-offs in the lowest end of the accuracy range. At the same time, models with $W=9$ are rarely on the frontier, highlighting at the same time the importance of processing a sequence of IR frames to achieve high accuracy, and the fact that too long sequences stop providing useful information and lead to over-fitting. 
Lastly, comparing Fig.~\ref{fig:all}a and Fig.~\ref{fig:all}b, shows that quantization does not cause relevant accuracy drops, and rather yields a Bal. Acc. \textit{increase} on most models, especially the lowest complexity ones. This is due to its well-known regularizing effect~\cite{Jacob2018}, which again helps to reduce overfitting.

Going more into the details of each chart, we note that each Pareto front is formed by a different combination of model types, showing that different architectures are preferable for optimizing the model size or the number of operations.
Specifically, when considering the Bal. Acc. versus MAC graphs, Single-frame and Multi-channel CNNs (crosses in the charts) occupy most of the Pareto front, for both float and 8-bit models. In contrast, when considering the number of parameters as a cost metric, the front is mainly composed of Majority voting (squares) and Concatenated CNNs (circles). 
This is expected, since for $W>1$, Multi-channel CNNs require additional MACs only in the first Conv layer, whereas all subsequent layers remain identical to the case of $W=1$. In contrast, Majority-voting and Concatenated CNNs repeat the execution of the entire network, or feature extractor, on each frame, which makes the total MACs grow almost linearly with $W$. Therefore, these models ``pay'' the Bal. Acc. benefits deriving from a larger $W$ with a much larger number of operations. Vice versa, since the weights used to process each IR frame are \textit{shared}, the cost increase in terms of model size is lower. Specifically, it is near-zero for Majority-voting CNNs, and limited to the final FC layers for Concatenated CNNs. Accordingly, when considering the parameters as a cost metric, these models are able to outperform multi-channel CNNs and reach the Pareto frontier.

Predictably, the most complex models (CNN-LSTM and CNN-TCN) appear in the high-accuracy part of the Pareto curves. Namely, the most accurate floating point model is a CNN-LSTM, reaching 82.7\% Bal. Acc. with $\approx$ 14k MACs and 2.38k parameters, whereas two CNN-TCN appear in the MACs-related Pareto front for quantized models, close to the top. However, in general, most instances of these two types of model suffer from over-fitting, achieving sub-optimal performance, while incurring a high cost in terms of MACs and parameters, as shown by the fact that they mostly occupy the right side of the complete charts. Overall, we can conclude that simple and efficient solutions to combine multiple IR frames (multi-channel, mode inference, and feature concatenation) are preferable for this relatively simple task and small dataset.

\begin{table}[t]
    \centering
     \caption{Summary of the characteristics of the considered DL models.\label{tab:summary}}
    {
    \begin{tabular}{|c|c|c|c|}
    \hline
     \textbf{Model}   & \textbf{Best For} & \textbf{Bal. Acc. Target} & \textbf{Max Input Win.}\\ 
     \hline 
     \hline 
     Single-frame & Latency/Energy & Low & 1\\ \hline
     Multi-frame & Latency/Energy & Mid & 3\\ \hline
     Majority & Memory & Whole Range & 7\\ \hline
     Concat & Memory & Mid & 7\\ \hline
     CNN-LSTM & Memory & High & 9\\ \hline
     CNN-TCN & Latency/Energy & High & 3\\ \hline
    \end{tabular}}
\end{table}

Table~\ref{tab:summary} reports a summary of all considered DL models, highlighting the features and requirements of each type based on our results. The table reports only qualitative trends, since the exact numerical results could change for different sensors or datasets. Specifically, for each model type, we summarize our Pareto analysis on both floating point and quantized implementations, reporting: i) whether a given model is most effective for memory reduction or for latency/energy reduction, depending on whether it is found more frequently on the parameters or MACs Pareto frontier respectively (\textit{Best For} column); ii) the accuracy range for which such model is preferable (\textit{Bal. Acc. Target} column), which also implicitly defines the corresponding resource range (memory or latency/energy); iii) The maximum IR frames window length that yields Pareto-optimal results for that model family (\textit{Max Input Win.}). Approximately, Low, Mid, and High Bal. Acc. ranges correspond to $<75\%$, $75\%-80\%$ and $>80\%$ respectively. The table provides, at a glance, a general guidance for system designers. For instance, it shows that single-frame CNNs are a good choice when the objective is to obtain a fast and energy-efficient inference, and very high accuracy is not required. Similarly, it shows that Majority voting is a very effective solution for memory reduction, across the whole accuracy range, or that CNN-LSTMs are the only models for which a window length $>7$ is useful for improving accuracy, etc.

\begin{table*}[t]
    \centering
     \caption{Detailed evaluation and deployment results of selected architectures.\label{tab:deploy}}
    \resizebox{\textwidth}{!}{
    \begin{tabular}{|c|c|c|c|c|c|c|c|c|c|c|}
    \hline
     \textbf{Model} &  \textbf{Bal. Acc.} & \textbf{Acc.}
      & \textbf{F1} & \textbf{MSE} & \textbf{MAE} &    \textbf{Model Size} & \textbf{Tot. Mem.}  &  \textbf{Energy} & \textbf{Latency} & \textbf{Architecture}  \\ 
                     &  \textbf{[\%]}      & \textbf{[\%]}             
      &             &              &              &    \textbf{[kB]} & \textbf{[kB]}  &  \textbf{[µJ]} & \textbf{[ms]} & \\  \hline \hline

    Top & 82.70±6.15  & 84.34±7.84 & 0.85±0.07 & 0.18±0.09 & 0.16±0.08 & 9.28 & 82.38 & 80.26±0.10 & 5.16±0.0064 & [$\blacktriangle$3] C8-P-C8-L16-FC \\

    Size-H & 76.25±5.54  & 78.13±9.08 & 0.79±0.08 & 0.24±0.09 & 0.23±0.09 & 2.97 & 68.73 & 54.96±0.01 & 3.53±0.0006 & [$\bullet$3] C8-P-C8-Cat-FC \\ 
    
        MAC-H & 77.62±5.98 & 78.04±8.18 &  0.80±0.07 & 0.27±0.11  & 0.24±0.09 & 5.7  & 42.95 & 29.25±0.01 & 1.88±0.0003 & [$\times$3] C8-P-C16-FC \\

     Size-L & 57.08±11.37 &51.10±22.49 & 0.52±0.23 &0.78±0.70 & 0.58±0.36 & 1.45   & 37.88 & 85.75±0.06 & 5.51±0.0036 & [$\blacksquare$5] C8-P-FC \\

   MAC-L  &  55.70±11.86 & 50.35±21.31 & 0.51±0.21 & 0.81±0.68 & 0.59±0.35 & 1.45 & 37.59 & 17.18±0.01 & 1.10±0.0007 & [$\times$1] C8-P-FC \\ \hline \hline

    Top-Q &  82.17±6.42 & 86.06±5.59 & 0.86±0.05 & 0.15±0.05 & 0.14±0.05 & 1.71  & 78.01 & 120.43±0.02 & 7.74±0.0010 & [$\blacksquare$5] C8-P-C8-FC-FC \\

    Size-H-Q/MAC-H-Q & 77.08±6.05 & 79.48±6.53 & 0.81±0.06 & 0.24±0.07 & 0.22±0.07 & 0.9 & 76.32 & 27.70±0.02 & 1.78±0.0010   & [$\times$3] C8-P-C8-FC  \\ 
    
     Size-L-Q & 63.87±10.76 & 70.83±13.79 & 0.70±0.14 & 0.33±0.12 & 0.30±0.13 & 0.41 & 71.56  & 61.90±0.02 & 3.98±0.0010 & [$\blacksquare$3] C8-P-FC \\ 
         
       MAC-L-Q & 62.88±7.52 & 68.97±14.03 & 0.69±0.14 & 0.36±0.13 & 0.33±0.14 & 0.41 & 71.39 &  20.45±0.01 & 1.32±0.0007 & [$\times$1] C8-P-FC \\ \hline \hline

   \cite{grideye}  & 42.77±14.50 & 57.54±11.50 & 0.56±0.12 &  0.61±0.21 & 0.49±0.14 & -  & 20.07 & 60.34±0.005 & 3.88±0.0003 & - \\

    \hline
    \end{tabular}}
\end{table*}

\subsection{Deployment}

We have selected 5 floating point and 5 quantized architectures from the Pareto curves derived in Sec.~\ref{sec:exploration} to deploy on the target MCU. Namely, we deployed: i) the model achieving the best balanced accuracy (Top); ii) the smallest model overall (Size-L) and the one requiring the least number of MACs (MAC-L); iii) the smallest/fewest-MAC models that achieve a Bal. Acc. drop $<$ 5\% with respect to Top (Size-H/MAC-H).

Table~\ref{tab:deploy} shows the detailed deployment results for these architectures on the STM32L4A6ZG MCU. Quantized models are denoted with a ``-Q'' suffix. Besides people counting accuracy metrics, we also report the memory occupation, energy consumption and inference latency of each model. In particular, for what concerns memory, we report both the model size and the total occupied Flash, which also includes code size. The same quantities are also reported for~\cite{grideye} for comparison. The rightmost column summarizes the architecture of each deployed neural network. Namely, the symbols inside the square brackets indicate the model type, using the same marker shape of Fig.~\ref{fig:all} (e.g., $\blacktriangle$ corresponds to a CNN-LSTM). The number in brackets corresponds to the value of $W$. Then, the sequence of layers in the model is encoded as follows: ``C$n$'' corresponds to a Conv layer with $n$ output channels, with implicit BatchNorm and ReLU, ``FC'' is a fully-connected layer, ``P'' a max. pooling layer, ``L$m$'' a LSTM cell with hidden size $m$, and ``Cat'' a concatenation.

As shown, all quantized models, as well as most floating-point models (except Size-L and MAC-L) greatly outperform \cite{grideye} in all considered accuracy metrics. In terms of balanced accuracy, our models outperform \cite{grideye} by 20.1\%-39.4\% and 12.9-39.9\% for integer and floating-point data representations respectively. Moreover, MAC-H and MAC-L in both implementations are faster and more energy efficient (from 2.06x to 3.51x) than \cite{grideye}, while still significantly outperforming it. For example, MAC-H-Q is 2.18x times faster and more energy efficient than \cite{grideye}, while also achieving $+34.3$\% Bal. Acc., $+21.9$\% Acc., 1.44x higher F1 Score, and 2.54x/2.22x lower MSE/MAE. 

The model size of all selected architectures is extremely small, with the smallest one occupying only 0.408 KB. The total memory, instead, is larger than \cite{grideye}, but this is mostly due to the large code size of X-CUBE-AI libraries, which contributes to up to 97\% of the Flash occupation. As shown in the table, the resulting memory depends on the types of layers present in the model (e.g., the ``Top'' floating point model requires more memory partly because of the additional inclusion of LSTM-related code). Further, quantized models have a larger code size compared to floating point ones on average, probably due to the more complex logic for handling scaling factors and re-quantization operations~\cite{Jacob2018}. Nonetheless, all considered models can easily fit in memory-limited IoT nodes, requiring 37.6-82.4kB of Flash, which corresponds to 3.7\%-8\% of the 1MB available on the MCU considered for our experiments.

All our models also have a latency $< 10$ms , which is below the real-time constraint, considering the 10FPS acquisition rate of our target dataset. Furthermore, considering a small 1400mAh@3.7V battery, and ignoring non-idealities and conversion losses for simplicity, a model such as MAC-H-Q would be able to continuously run inferences at that frame rate for more than 2 years without recharging.

\subsection{Comparison with state-of-the-art ML/DL Approaches}

Table \ref{tab:comparison} compares our work with the most relevant Machine Learning and Deep Learning approaches for people counting with IR array sensors. Of course, the comparison is only qualitative, since most previous works have been tested on private datasets, and deployed on different hardware. In the table, besides the input frame size, we report the Acc., F1 and MSE scores when available (other metrics were not considered by previous works). All scores are directly taken from the original papers. We also report the model size and the number of operations (OPs) per inference, as two hardware-independent complexity metrics. For DL solutions, we approximate OPs with the number of MACs, and when either Size or OPs are not reported by the authors, we calculated them based on the layers' geometries. For~\cite{chidurala2021occupancy}, instead, estimating Size and OPs was not possible, since the authors did not report the number and maximum depth of the decision trees that compose their best-performing RF.

\begin{table}[t]
    \centering
     \caption{Comparison with the State of the Art.\label{tab:comparison}}
    \begin{threeparttable}
    \begin{tabular}{|l|l|l|l|l|l|l|l|}
    \hline
    \textbf{Result} & \textbf{Input} & \textbf{Acc. [\%]} & \textbf{F1} & \textbf{MSE} & \textbf{Size [kB]} & \textbf{OPs}\\
    \hline
    \hline
    \cite{bouazizi2022low}                 & 8x6   & n.a.         & 0.88          & n.a.          & 450           & 34$\cdot$10\textsuperscript{6}\\
    \cite{gomez2018thermal}$^{*,\dagger}$  & 10x10 & 95.9         & n.a.          & n.a.          & 13.6          & 117$\cdot$10\textsuperscript{3}\\
    \cite{metwaly2019edge}$^{*}$           & 32x24 & \textbf{98.9}& n.a.          & \textbf{0.01} & 400           & 400$\cdot$10\textsuperscript{3}\\
    \cite{chidurala2021occupancy}$^{*}$    & 8x8   & 94.6         & 0.95          & n.a.          & n.a.          & n.a.\\
    \cite{en14154542}$^{\S}$   & 32x24   & 94.1       & n.a.         & 0.057       & 520.8        & 25$\cdot$10\textsuperscript{6} \\\hline
    \hline        
    \textbf{Top-Q}                         & 8x8   & 86.1         & 0.86          & 0.15          & \textbf{1.71} & \textbf{20$\cdot$10\textsuperscript{3}}\\
    \textbf{Top-Q}$^{*}$                   & 8x8   & 95.3         & \textbf{0.95} & 0.05          & 20.1          & 80$\cdot$10\textsuperscript{3}\\\hline
    \end{tabular}
    \begin{tablenotes}
    \item[(*)] Train/test split based on random sampling, not per-session.
    
    \item[($\S$)] Train/validation/test split based on sequences splitting in different locations, not per-session.
    
    \item[($\dagger$)] Numbers refer to the processing of a single 10x10 sliding window. This approach also \textit{localizes} people.

    \end{tablenotes}
    \end{threeparttable}
\end{table}

We report two results for our work: the first one corresponds to the ``Top-Q'' network of Table~\ref{tab:deploy}, found using the described per-session CV approach. Additionally, since~\cite{gomez2018thermal,metwaly2019edge,chidurala2021occupancy} use a purely random sampling method to separate training and test sets, we also report the best quantized results obtained with such kind of splitting. Precisely, we repeat the architecture search using a random 80\%/20\% train/test split, and report the average test set results over 4 iterations. Note that the resulting Top-Q$^*$ model has a different architecture from Top-Q. Namely, it is a quantized CNN-TCN model with the following structure: [$\blacklozenge$9] C8-P-C32-TCN32-FC-FC, where TCN$o$ refers to a TCN (1D Conv) layer with $o$ output channels.

Our main reference for comparison among state-of-the-art DL methods is~\cite{bouazizi2022low}, which uses a per-session split and a similar input resolution. Compared to this work, we obtain a comparable F1 Score with our Top-Q, but since our classification model is significantly smaller, and we do not need an additional super-resolution network, we achieve a 263x reduction in size and 1700x fewer OPs. The work of~\cite{en14154542} also uses a time-based data split, although simpler than ours: they assign to different data buckets the frames collected in the same location at different times. Their work achieves a higher Accuracy than our Top-Q (94.1\% vs 86.1\%) but this is mostly due to the 12x higher resolution input. Furthermore, their model requires about 130k floating point parameters, resulting in a model size of 521kB, which is 304x more than that of Top-Q. Similarly, the number of OPs is in the order of millions, more than 1000x larger than Top-Q.

Since the dataset of \cite{en14154542} is publicly available, we also ran two additional experiments on it. First, we down-sampled the images to 8x8 resolution and excluded all samples with more than 3 people to fairly compare with LINAIGE. Then, we trained our ``Top'' model from Table~\ref{tab:deploy} using only the data from~\cite{en14154542} and maintaining our training protocol. We obtained an accuracy of 72.6\%, much lower than the one achieved by their model, but acceptable given the lower resolution of our inputs and the striking $>$ 1000x complexity reduction. Further, the dataset in~\cite{en14154542} only contains $\approx$ 9k samples with less than 3 people, versus the $>$ 20k of LINAIGE. Thus, we also tried to use the down-sampled data from \cite{en14154542} to \textit{augment} the LINAIGE training dataset in each CV fold. In this case, the ``Top'' model improves in all classification metrics on average (Acc. +4.8\%, F1 +0.04, MSE -0.06, MAE -0.05) except for the Bal. Acc (-1.2\%) with respect to pure LINAIGE training. This shows that, potentially, using a larger dataset could further improve the results achieved by our efficient DL models, especially the most complex architectures. 

When considering a random data split, Top-Q$^*$ obtains slightly lower accuracy and higher MSE compared to~\cite{metwaly2019edge}, but uses a smaller-resolution input, and requires a 233x smaller model and 20x fewer inference operations. It also achieves comparable accuracy and F1 score with respect to the RF-based approach of~\cite{chidurala2021occupancy}. Lastly, \cite{gomez2018thermal} uses a model smaller than Top-Q$^*$ to achieve a slightly higher accuracy on a 10x10 input.  However, the inputs processed by~\cite{gomez2018thermal} are patches extracted from a much higher resolution input (80x60), which is further upscaled to 120x90 and 160x120. All three versions of the image are then processed by the CNN in 10x10 sliding-window patches. Therefore, the total number of inference operations is huge for this solution (approximately 450$\cdot$10\textsuperscript{6} based on our calculations), which translates into very long latencies and high energy consumption. Indeed, the authors report a total latency of 63s and an energy of 2.2J, orders of magnitude higher than those achieved by our models. It must be underlined that~\cite{gomez2018thermal} attempts not only to count people in the frame, but also to localize their heads, which is significantly different from our goal, and only possible due to their higher-resolution input. Indeed, the 95.9\% accuracy reported in the table refers to head detection on a single 10x10 patch, whereas the final counting accuracy is just 53.7\%. 

In summary, these comparisons show that our proposed models achieve comparable counting accuracy with much lower complexity on average, compared to state-of-the-art solutions. This is particularly important for deployment at the IoT edge, where devices have very tight memory budgets, and extreme constraints in terms of energy consumption, being typically battery powered and expected to operate for years without recharging. The tiny and efficient DL models explored in this work could enable novel pervasive and privacy-preserving people counting solutions in environments where access to the power grid is not available, which would exclude most of the energy-hungry state-of-the-art solutions.

%% file: sec/conclusion.tex
We have conducted the first systematic study on efficient DL architectures for person counting based on ultra-low-resolution IR arrays, obtaining a wide range of trade-offs between classification scores, memory occupation, latency and energy consumption, and showing that different types of DL models are preferable for different objectives. The resulting Pareto-optimal models obtain counting accuracy scores that are significantly higher than those of a publicly available deterministic solution~\cite{grideye} (up to 82.70\% balanced accuracy vs 42.77\%), and comparable with a state-of-the-art DL approach~\cite{bouazizi2022low} (0.86 vs 0.88 F1-score), while reducing the latency and energy requirements by up to more than 2x with respect to the former, e.g. 1.78ms/27.70$\mu$J vs 3.88ms/60.34$\mu$J per inference at approximately $+34.3\%$ balanced accuracy for our method. Furthermore, our models enable continuous real-time inference ($<10$ms latency) on IoT edge devices based on MCUs, with years of autonomous operation, while requiring less than 100kB of memory.

%% file: main.bbl
\begin{thebibliography}{10}
\providecommand{\url}[1]{#1}
\csname url@samestyle\endcsname
\providecommand{\newblock}{\relax}
\providecommand{\bibinfo}[2]{#2}
\providecommand{\BIBentrySTDinterwordspacing}{\spaceskip=0pt\relax}
\providecommand{\BIBentryALTinterwordstretchfactor}{4}
\providecommand{\BIBentryALTinterwordspacing}{\spaceskip=\fontdimen2\font plus
\BIBentryALTinterwordstretchfactor\fontdimen3\font minus
  \fontdimen4\font\relax}
\providecommand{\BIBforeignlanguage}[2]{{%
\expandafter\ifx\csname l@#1\endcsname\relax
\typeout{** WARNING: IEEEtran.bst: No hyphenation pattern has been}%
\typeout{** loaded for the language `#1'. Using the pattern for}%
\typeout{** the default language instead.}%
\else
\language=\csname l@#1\endcsname
\fi
#2}}
\providecommand{\BIBdecl}{\relax}
\BIBdecl

\bibitem{chen2019deep}
J.~Chen \emph{et~al.}, ``Deep learning with edge computing: A review,''
  \emph{Proceedings of the IEEE}, vol. 107, no.~8, pp. 1655--1674, 2019.

\bibitem{Jiang2019}
B.~Jiang \emph{et~al.}, ``Wearable vision assistance system based on binocular
  sensors for visually impaired users,'' \emph{IEEE Internet of Things
  Journal}, vol.~6, no.~2, pp. 1375--1383, 2019.

\bibitem{Khan2020}
K.~Muhammad \emph{et~al.}, ``Cost-effective video summarization using deep cnn
  with hierarchical weighted fusion for iot surveillance networks,'' \emph{IEEE
  Internet of Things Journal}, vol.~7, no.~5, pp. 4455--4463, 2020.

\bibitem{Burrello2022a}
A.~Burrello \emph{et~al.}, ``Bioformers: Embedding transformers for ultra-low
  power semg-based gesture recognition,'' in \emph{2022 Design, Automation \&
  Test in Europe Conference \& Exhibition (DATE)}, 2022, pp. 1443--1448.

\bibitem{Risso2022}
M.~Risso \emph{et~al.}, ``Lightweight neural architecture search for temporal
  convolutional networks at the edge,'' \emph{IEEE Transactions on Computers},
  pp. 1--1, 2022.

\bibitem{Zhou2019}
Z.~Zhou \emph{et~al.}, ``{Edge Intelligence: Paving the Last Mile of Artificial
  Intelligence With Edge Computing},'' \emph{Proceedings of the IEEE}, vol.
  107, no.~8, pp. 1738--1762, 2019.

\bibitem{Shi2016}
W.~Shi \emph{et~al.}, ``{Edge Computing: Vision and Challenges},'' \emph{IEEE
  Internet of Things Journal}, vol.~3, no.~5, pp. 637--646, 2016.

\bibitem{hou2010people}
Y.-L. Hou \emph{et~al.}, ``People counting and human detection in a challenging
  situation,'' \emph{IEEE transactions on systems, man, and cybernetics-part a:
  systems and humans}, vol.~41, no.~1, pp. 24--33, 2010.

\bibitem{tsou2020counting}
P.-R. Tsou \emph{et~al.}, ``Counting people by using convolutional neural
  network and a pir array,'' in \emph{2020 21st IEEE International Conference
  on Mobile Data Management (MDM)}.\hskip 1em plus 0.5em minus 0.4em\relax
  IEEE, 2020, pp. 342--347.

\bibitem{xie2022privacy}
C.~Xie \emph{et~al.}, ``Privacy-preserving social distance monitoring on
  microcontrollers with low-resolution infrared sensors and cnns,'' in
  \emph{Proceedings of the 2022 IEEE International Symposium on Circuits and
  Systems (ISCAS)}, ser. ISCAS 2022.\hskip 1em plus 0.5em minus 0.4em\relax
  IEEE, 2022.

\bibitem{perra2021monitoring}
C.~Perra \emph{et~al.}, ``Monitoring indoor people presence in buildings using
  low-cost infrared sensor array in doorways,'' \emph{Sensors}, vol.~21,
  no.~12, p. 4062, 2021.

\bibitem{raghavachari2015comparative}
C.~Raghavachari \emph{et~al.}, ``A comparative study of vision based human
  detection techniques in people counting applications,'' \emph{Procedia
  Computer Science}, vol.~58, pp. 461--469, 2015.

\bibitem{xi2014electronic}
W.~Xi \emph{et~al.}, ``Electronic frog eye: Counting crowd using wifi,'' in
  \emph{IEEE INFOCOM 2014-IEEE Conference on Computer Communications}.\hskip
  1em plus 0.5em minus 0.4em\relax IEEE, 2014, pp. 361--369.

\bibitem{hashimoto1997people}
K.~Hashimoto \emph{et~al.}, ``People count system using multi-sensing
  application,'' in \emph{Proceedings of International Solid State Sensors and
  Actuators Conference (Transducers' 97)}, vol.~2.\hskip 1em plus 0.5em minus
  0.4em\relax IEEE, 1997, pp. 1291--1294.

\bibitem{udrea2021new}
I.~Udrea \emph{et~al.}, ``New research on people counting and human
  detection,'' in \emph{2021 13th International Conference on Electronics,
  Computers and Artificial Intelligence (ECAI)}.\hskip 1em plus 0.5em minus
  0.4em\relax IEEE, 2021, pp. 1--6.

\bibitem{shami2018people}
M.~B. Shami \emph{et~al.}, ``People counting in dense crowd images using sparse
  head detections,'' \emph{IEEE Transactions on Circuits and Systems for Video
  Technology}, vol.~29, no.~9, pp. 2627--2636, 2018.

\bibitem{shetty2017detection}
A.~D. Shetty \emph{et~al.}, ``Detection and tracking of a human using the
  infrared thermopile array sensor—“grid-eye”,'' in \emph{2017
  International Conference on Intelligent Computing, Instrumentation and
  Control Technologies (ICICICT)}.\hskip 1em plus 0.5em minus 0.4em\relax IEEE,
  2017, pp. 1490--1495.

\bibitem{basalamah2019scale}
S.~Basalamah \emph{et~al.}, ``Scale driven convolutional neural network model
  for people counting and localization in crowd scenes,'' \emph{IEEE Access},
  vol.~7, pp. 71\,576--71\,584, 2019.

\bibitem{nogueira2019retailnet}
V.~Nogueira \emph{et~al.}, ``Retailnet: A deep learning approach for people
  counting and hot spots detection in retail stores,'' in \emph{2019 32nd
  SIBGRAPI Conference on Graphics, Patterns and Images (SIBGRAPI)}.\hskip 1em
  plus 0.5em minus 0.4em\relax IEEE, 2019, pp. 155--162.

\bibitem{khan2019person}
S.~D. Khan \emph{et~al.}, ``Person head detection based deep model for people
  counting in sports videos,'' in \emph{2019 16th IEEE International Conference
  on Advanced Video and Signal Based Surveillance (AVSS)}.\hskip 1em plus 0.5em
  minus 0.4em\relax IEEE, 2019, pp. 1--8.

\bibitem{bouazizi2022low}
M.~Bouazizi \emph{et~al.}, ``Low-resolution infrared array sensor for counting
  and localizing people indoors: When low end technology meets cutting edge
  deep learning techniques,'' \emph{Information}, vol.~13, no.~3, p. 132, 2022.

\bibitem{gomez2018thermal}
A.~Gomez \emph{et~al.}, ``Thermal image-based cnn's for ultra-low power people
  recognition,'' in \emph{Proceedings of the 15th ACM International Conference
  on Computing Frontiers}, 2018, pp. 326--331.

\bibitem{metwaly2019edge}
A.~Metwaly \emph{et~al.}, ``Edge computing with embedded ai: Thermal image
  analysis for occupancy estimation in intelligent buildings,'' in
  \emph{Proceedings of the INTelligent Embedded Systems Architectures and
  Applications Workshop 2019}, 2019, pp. 1--6.

\bibitem{en14154542}
\BIBentryALTinterwordspacing
M.~Kraft \emph{et~al.}, ``Low-cost thermal camera-based counting occupancy
  meter facilitating energy saving in smart buildings,'' \emph{Energies},
  vol.~14, no.~15, 2021. [Online]. Available:
  \url{https://www.mdpi.com/1996-1073/14/15/4542}
\BIBentrySTDinterwordspacing

\bibitem{grideye}
P.~Industry, ``Grid-eye application note on social distancing. people detection
  and tracking with ceiling mounted sensors,'' 2020.

\bibitem{mohammadmoradi2017measuring}
H.~Mohammadmoradi \emph{et~al.}, ``Measuring people-flow through doorways using
  easy-to-install ir array sensors,'' in \emph{2017 13th International
  Conference on Distributed Computing in Sensor Systems (DCOSS)}.\hskip 1em
  plus 0.5em minus 0.4em\relax IEEE, 2017, pp. 35--43.

\bibitem{wang2021lightweight}
H.~Wang \emph{et~al.}, ``A lightweight people counting approach for smart
  buildings,'' in \emph{2021 13th International Conference on Wireless
  Communications and Signal Processing (WCSP)}.\hskip 1em plus 0.5em minus
  0.4em\relax IEEE, 2021, pp. 1--5.

\bibitem{rabiee2021multi}
R.~Rabiee \emph{et~al.}, ``Multi-bernoulli tracking approach for occupancy
  monitoring of smart buildings using low-resolution infrared sensor array,''
  \emph{Remote Sensing}, vol.~13, no.~16, p. 3127, 2021.

\bibitem{kawashima2017}
T.~Kawashima \emph{et~al.}, ``Action recognition from extremely low-resolution
  thermal image sequence,'' in \emph{Proceedings of the 14th IEEE International
  Conference on Advanced Video and Signal Based Surveillance}, Aug. 2017, pp.
  1--6.

\bibitem{singh2019non}
S.~Singh \emph{et~al.}, ``Non-intrusive presence detection and position
  tracking for multiple people using low-resolution thermal sensors,''
  \emph{Journal of Sensor and Actuator Networks}, vol.~8, no.~3, p.~40, 2019.

\bibitem{linaige}
\BIBentryALTinterwordspacing
C.~Xie \emph{et~al.}, ``Low-resolution infrared-array data for ai on the
  edge,'' 2022. [Online]. Available:
  \url{https://www.kaggle.com/datasets/francescodaghero/linaige}
\BIBentrySTDinterwordspacing

\bibitem{chidurala2021occupancy}
V.~Chidurala \emph{et~al.}, ``Occupancy estimation using thermal imaging
  sensors and machine learning algorithms,'' \emph{IEEE Sensors Journal},
  vol.~21, no.~6, pp. 8627--8638, 2021.

\bibitem{xieEnergyefficient2022}
C.~Xie \emph{et~al.}, ``Energy-efficient and {{Privacy-aware Social Distance
  Monitoring}} with {{Low-resolution Infrared Sensors}} and {{Adaptive
  Inference}},'' in \emph{2022 17th {{Conference}} on {{Ph}}.{{D Research}} in
  {{Microelectronics}} and {{Electronics}} ({{PRIME}})}, Jun. 2022, pp.
  181--184.

\bibitem{liu2005detecting}
X.~Liu \emph{et~al.}, ``Detecting and counting people in surveillance
  applications,'' in \emph{IEEE Conference on Advanced Video and Signal Based
  Surveillance, 2005.}\hskip 1em plus 0.5em minus 0.4em\relax IEEE, 2005, pp.
  306--311.

\bibitem{stec2019multi}
M.~Stec \emph{et~al.}, ``Multi-sensor-fusion system for people counting
  applications,'' in \emph{2019 First International Conference on Societal
  Automation (SA)}.\hskip 1em plus 0.5em minus 0.4em\relax IEEE, 2019, pp.
  1--4.

\bibitem{olmeda2013pedestrian}
D.~Olmeda \emph{et~al.}, ``Pedestrian detection in far infrared images,''
  \emph{Integrated Computer-Aided Engineering}, vol.~20, no.~4, pp. 347--360,
  2013.

\bibitem{flir}
FLIR, ``Free flir thermal dataset for algorithm training,'' 2018.

\bibitem{hwang2015multispectral}
S.~Hwang \emph{et~al.}, ``Multispectral pedestrian detection: Benchmark dataset
  and baselines,'' in \emph{Proceedings of IEEE Conference on Computer Vision
  and Pattern Recognition (CVPR)}, 2015.

\bibitem{Rivadeneira_2020_CVPR_Workshops}
R.~E. Rivadeneira \emph{et~al.}, ``Thermal image super-resolution challenge -
  pbvs 2020,'' in \emph{Proceedings of the IEEE/CVF Conference on Computer
  Vision and Pattern Recognition (CVPR) Workshops}, June 2020.

\bibitem{baja-1j59-20}
\BIBentryALTinterwordspacing
Y.~Karayaneva \emph{et~al.}, ``Infrared human activity recognition dataset -
  coventry-2018,'' 2020. [Online]. Available:
  \url{https://dx.doi.org/10.21227/baja-1j59}
\BIBentrySTDinterwordspacing

\bibitem{He2017}
\BIBentryALTinterwordspacing
K.~He \emph{et~al.}, ``Mask r-cnn,'' 2017. [Online]. Available:
  \url{https://arxiv.org/abs/1703.06870}
\BIBentrySTDinterwordspacing

\bibitem{lam1997application}
L.~Lam \emph{et~al.}, ``Application of majority voting to pattern recognition:
  an analysis of its behavior and performance,'' \emph{IEEE Transactions on
  Systems, Man, and Cybernetics-Part A: Systems and Humans}, vol.~27, no.~5,
  pp. 553--568, 1997.

\bibitem{lee1993handprinted}
D.-S. Lee, ``Handprinted digit recognition: A comparison of algorithms,'' in
  \emph{Proceedings of the Third International Workshop on Frontiers in
  Handwriting Recognition}, 1993, pp. 153--164.

\bibitem{amin2020cnns}
M.~Amin-Naji \emph{et~al.}, ``Cnns hard voting for multi-focus image fusion,''
  \emph{Journal of Ambient Intelligence and Humanized Computing}, vol.~11,
  no.~4, pp. 1749--1769, 2020.

\bibitem{Yazdizadeh2020}
A.~Yazdizadeh \emph{et~al.}, ``Ensemble convolutional neural networks for mode
  inference in smartphone travel survey,'' \emph{IEEE Transactions on
  Intelligent Transportation Systems}, vol.~21, no.~6, pp. 2232--2239, 2020.

\bibitem{demir2020new}
F.~Demir \emph{et~al.}, ``A new pyramidal concatenated cnn approach for
  environmental sound classification,'' \emph{Applied Acoustics}, vol. 170, p.
  107520, 2020.

\bibitem{wu2019concatenate}
Q.~Wu \emph{et~al.}, ``Concatenate convolutional neural networks for
  non-intrusive load monitoring across complex background,'' \emph{Energies},
  vol.~12, no.~8, p. 1572, 2019.

\bibitem{kim2019predicting}
T.-Y. Kim \emph{et~al.}, ``Predicting residential energy consumption using
  cnn-lstm neural networks,'' \emph{Energy}, vol. 182, pp. 72--81, 2019.

\bibitem{zhao2019speech}
J.~Zhao \emph{et~al.}, ``Speech emotion recognition using deep 1d \& 2d cnn
  lstm networks,'' \emph{Biomedical signal processing and control}, vol.~47,
  pp. 312--323, 2019.

\bibitem{sciannameoDeep2022}
V.~Sciannameo \emph{et~al.}, ``A deep learning approach for {{Spatio-Temporal}}
  forecasting of new cases and new hospital admissions of {{COVID-19}} spread
  in {{Reggio Emilia}}, {{Northern Italy}},'' \emph{Journal of Biomedical
  Informatics}, vol. 132, p. 104132, Aug. 2022.

\bibitem{leaTemporal2016}
C.~Lea \emph{et~al.}, ``Temporal {{Convolutional Networks}}: {{A Unified
  Approach}} to {{Action Segmentation}},'' in \emph{Computer {{Vision}}
  \textendash{} {{ECCV}} 2016 {{Workshops}}}, ser. Lecture {{Notes}} in
  {{Computer Science}}, G.~Hua \emph{et~al.}, Eds.\hskip 1em plus 0.5em minus
  0.4em\relax {Cham}: {Springer International Publishing}, 2016, pp. 47--54.

\bibitem{Burrello2021c}
A.~Burrello \emph{et~al.}, ``{Q-PPG: Energy-Efficient PPG-based Heart Rate
  Monitoring on Wearable Devices},'' \emph{IEEE Transactions on Biomedical
  Circuits and Systems}, p.~1, 2021.

\bibitem{tensorflow}
\BIBentryALTinterwordspacing
M.~Abadi \emph{et~al.}, ``Tesorflow: Large-scale machine learning on
  heterogeneous systems,'' 2015. [Online]. Available:
  \url{https://www.tensorflow.org/}
\BIBentrySTDinterwordspacing

\bibitem{Daghero2021a}
F.~Daghero \emph{et~al.}, ``{Energy-efficient deep learning inference on edge
  devices},'' in \emph{Hardware Accelerator Systems for Artificial Intelligence
  and Machine Learning}, ser. Advances in Computers, S.~Kim \emph{et~al.},
  Eds.\hskip 1em plus 0.5em minus 0.4em\relax Elsevier, 2021, vol. 122, ch.~8,
  pp. 247--301.

\bibitem{Jacob2018}
B.~Jacob \emph{et~al.}, ``{Quantization and Training of Neural Networks for
  Efficient Integer-Arithmetic-Only Inference},'' in \emph{Proceedings of the
  IEEE Conference on Computer Vision and Pattern Recognition (CVPR)}, jun 2018.

\bibitem{cubeai}
STMicroelectronics, ``{X-CUBE-AI, AI expansion pack for STM32CubeMX},''
  https://www.st.com/en/embedded-software/x-cube-ai.html.

\bibitem{mcu}
------, ``{STM32L4A6ZG, Ultra-low-power Arm Cortex-M4 32-bit MCU},''
  https://www.st.com/en/microcontrollers-microprocessors/stm32l4a6zg.html.

\bibitem{stcubeide}
------, ``{STM32CubeIDE, Integrated Development Environment for STM32},''
  https://www.st.com/en/development-tools/stm32cubeide.htm.

\end{thebibliography}
